\documentclass[times,twocolumn,final,authoryear]{elsarticle}

\usepackage{prletters}
\usepackage{framed,multirow}

\usepackage{amssymb}
\usepackage{latexsym}

\usepackage[cmex10]{amsmath}
\usepackage{array}
\usepackage[lofdepth,lotdepth]{subfig}
\usepackage{epsfig}

\usepackage{url}
\usepackage{xcolor}
\definecolor{newcolor}{rgb}{.8,.349,.1}

\journal{Pattern Recognition Letters}

\begin{document}

\begin{frontmatter}

\title{Multi-hypothesis contextual modeling for semantic segmentation }

\author[1]{Hasan F. \snm{Ates}\corref{cor1}}
\cortext[cor1]{Corresponding author:
  Tel.: +90-216-681 5100 (x5643)}
\ead{hfates@medipol.edu.tr}
\author[2]{Sercan \snm{Sunetci}}
\address[1]{Dept. of Computer Engineering, Istanbul Medipol University, Istanbul, 34810 TURKEY}
\address[2]{Dept. of Electrical \& Electronics Engineering,  Isik University, Istanbul, 34980 TURKEY}


\begin{abstract}
Semantic segmentation (i.e. image parsing) aims to annotate each image pixel with its corresponding semantic class label. Spatially consistent labeling of the image requires an accurate description and modeling of the local contextual information. Segmentation result is typically improved by Markov Random Field (MRF) optimization on the initial labels. However this improvement is limited by the accuracy of initial result and how the contextual neighborhood is defined. In this paper, we develop generalized and flexible contextual models for segmentation neighborhoods in order to improve parsing accuracy. Instead of using a fixed segmentation and neighborhood definition, we explore various contextual models for fusion of complementary information available in alternative segmentations of the same image. In other words, we propose a novel MRF framework that describes and optimizes the contextual dependencies between multiple segmentations. Simulation results on two common datasets demonstrate significant improvement in parsing accuracy over the baseline approaches.
\end{abstract}


\end{frontmatter}


\section{Introduction}
\vspace{-0.1cm}
Semantic segmentation (i.e. image parsing) is a fundamental problem in computer vision. The goal is to segment the image accurately and annotate each segment with its true semantic class. Recent literature has seen two major trends in this problem. Superpixel-based segmentation and parsing algorithms (\cite{superparsing,fischer,Tighe2015}) are able to achieve much higher accuracy than similar pixel-based approaches. In superpixel-based segmentation, the image is segmented into visually meaningful atomic regions that agree with  object boundaries. Then the parsing algorithm assigns the same semantic label to all the pixels of a superpixel, resulting in a spatially smooth labeling of the whole image. However, with the advent of deep networks in machine learning, state-of-the-art accuracy is obtained by dense labeling of image pixels through the use of convolutional neural network (CNN) architectures  (\cite{fcnn,parsenet,rcnn}). Convolutional nets provide spatially dense but smooth classification by utilizing multiple pooling and upsampling layers (\cite{fcnn}).

\makeatletter{\renewcommand*{\@makefnmark}{}
\footnotetext{\copyright 2018. This manuscript version is made available under the CC-BY-NC-ND 4.0 license http://creativecommons.org/licenses/by-nc-nd/4.0/}\makeatother}

Detailed but spatially consistent labeling is essential for accurate segmentation of the image. In order to improve parsing consistency and accuracy, superpixel-based methods typically incorporate Markov Random Field (MRF) modeling and inference in superpixel neighborhoods (\cite{adapt_super,context_rare}). Conditional Random Fields (CRFs) are also integrated in deep networks for further improvement (\cite{deeplab,crf_rcnn}). However this improvement is limited by the accuracy of initial result and how the contextual neighborhood is defined. In this paper, we claim that
committing to a single segmentation method and fixed neighborhood definition is rather restrictive for describing the rich contextual information in an image. Several works have shown before that parsing performance benefits from the fusion of multiple segmentations (\cite{multi_hypothesis,vieux}) and/or object detectors (\cite{Tighe2015,nguyen_cnn}). Nevertheless, these approaches fail to address the problem of contextual inference among alternative descriptions.



In this paper we extend our previous work in (\cite{Ates2017}) and investigate various adaptive contextual models to combine complementary information available in alternative scene segmentations.  These models incorporate not only the spatial neighborhood of adjacent segments/superpixels within the same segmentation but also the neighborhood of intersecting superpixels from different segmentations. We claim that these segmentations provide complementary information about the underlying object classes in the image. The proposed approach provides a unified framework to fuse the information obtained from multiple segmentations. In other words, MRF optimization is used to code contextual constraints both in the spatial and inter-segmentation neighborhoods for more consistent labeling. As a result, a more flexible description of neighborhood context is achieved, when compared to the fixed context of a single segmentation. In particular, this paper provides contributions in the following two aspects:
\begin{itemize}
\item  Alternative labelings of the same image are produced by using multiple segmentation methods and/or parameter/feature settings. Tested parsing methods include superpixel-based SuperParsing (\cite{superparsing}),  Fully Convolutional Neural Network (FCN) (\cite{fcnn}) and Pyramid Scene Parsing Network (PSP) (\cite{psp}).
\item A generalized model of local context is proposed to encode contextual constraints between alternatives and to fuse complementary information available in different segmentations.
\end{itemize}
The proposed MRF models are tested and compared on SIFT Flow (\cite{siftflow}) and 19-class subset of LabelMe (\cite{labelmesub}) datasets. Simulation results demonstrate significant improvement in parsing accuracy over the baseline approaches. When compared to (\cite{Ates2017}), this paper provides substantially better results, due to use of both CNN features and FCN/PSP networks. Parsing accuracies are also better than or competitive with the state-of-the-art in semantic scene segmentation.

In the next section we review the related work in contextual modeling for image parsing and contrast our approach with existing models. Section \ref{context} develops the general framework of the proposed MRF contextual model. Section \ref{model} gives the details of the model and the parsing algorithms. Section \ref{benzetim} provides and discusses the simulations results. Section \ref{sonuc} concludes the paper with ideas for future work.

\section{Related Work}
\label{related}
Context in image parsing is typically introduced in the form of MRF or CRF models that describe the local and/or global dependencies among object labels and scene content. Several CNN-based parsing methods adopt CRFs as a post-processing step to refine their outputs (\cite{crf1, crf2}). \cite{deeplab} employs a fully connected CRF among pixels to capture both local and global context. These methods require separate training steps for learning the CNN and CRF. Recurrent neural networks (RNNs) are also used to model context among pixels/objects (\cite{rnn1, rnn2}), hence introducing context information into the neural network architecture. \cite{crf_rcnn} shows how to formulate CRF model as an RNN; in this manner CRF can be combined with any CNN-based parser for end-to-end training of the whole network.

There exist multi-scale approaches that model multi-scale context as well  (\cite{multiscale_parse, ms_eigen, parsenet, psp}). \cite{multiscale_parse} uses a multiscale set of segmentations, including superpixels, to train a deep network, learn hierarchical features and find an optimal cover of the image out of many segmentations. In the end, this method also commits to a final fixed segmentation, which is claimed to be optimal,  but does not consider a joint optimization of alternative representations. \cite{ms_eigen} progressively refine its network output using a sequence of scales to provide dense labeling. FCN combines coarse layer features with fine, low-layer features for fusion of contextual information at different resolutions. \cite{parsenet} and \cite{psp} use pooling of local features at different scales to capture global context. However network layers are generated by rectangular convolution and regular downsampling, which does not comply with the actual geometry of the objects in the scene.

In spite of the success of CRFs and multi-scale models for semantic scene segmentation, these approaches are computationally costly both for learning and inference. In this paper, we focus on context models that require minimal training and that can be optimized efficiently. In literature, several superpixel-based parsing algorithms use MRF-based post-processing to smooth out superpixel labels and improve labeling consistency among neighboring superpixels (\cite{superparsing, adapt_super}). Then MRF inference is achieved with fast and effective min-cut/max-flow optimization algorithm. However, typically, these parsing algorithms commit to a single pre-segmentation of the image, which is not always consistent with the boundaries  of object classes.

To circumvent the shortcomings of previous models, our generalized MRF model defines a flexible framework to combine information coming from multiple segmentations and parsing methods. The closest work to our proposal is Associative Hierarchical Random Fields (AHRF) of \cite{ahrf}. AHRF provides a hierarchical MRF model for multiple segmentations at different scales. AHRF is introduced as a generalization of different MRF models defined over pixels, superpixels or a hierarchy of segmentations (such as \cite{pantofaru}). While AHRF defines a strict hierarchy between pixels, segments and super-segments, our model allows for combination of different segmentations without any fixed parent-child or coarse-fine scale relationship in between. In addition, we investigate the fusion of decisions from different (superpixel-based and CNN-based) parsers, while \cite{ahrf} does not explain how to extend AHRF to incorporate several different classifiers.

The main novelty of this paper is the fusion of multiple parsing methods within MRF formalism. \cite{vieux} also labels segments by late fusion of SVM classifiers over multiple segmentations; however, fusion is simply performed by taking the mean/max/multiplication of classifier probabilities in intersecting regions and label smoothing by relaxation labeling is treated as a post-processing step on the fused result. Methods such as \cite{dong}, \cite{urtasun}, \cite{morales} define hierarchical MRF models over multiple segmentations but do not consider segmentations and class scores coming from alternative methods. In these approaches, since the segmentations and their unary potentials at different levels of the hierarchy are not independently generated, there will be no significant complementary information for fusion over the hierarchical MRF. As a result, gains in labeling accuracy are limited. On the other hand, our MRF framework allows for the fusion of independent segmentations and class likelihoods coming from much different classifiers.

\section{Contextual Modeling of Alternative Segmentations}
\label{context}
In superpixel image parsing, a fixed segmentation is typically used to derive local features, estimate class likelihoods, label each segment and perform MRF smoothing of labels. Hence parsing performance heavily depends on how well the size, shape, boundary and content of  superpixels represent the underlying object classes.

%

In this paper we further develop ``Multi-hypothesis MRF model'' of \cite{Ates2017} and show that labeling accuracy could benefit from the joint use of multiple initial segmentations, which are possibly generated by different methods. In our approach the local context  incorporates not only the spatial neighborhood of adjacent superpixels within the same segmentation but also the neighborhood of intersecting superpixels from different segmentations. These inter-segmentation neighborhoods help fuse alternative representations coming from multiple segmentations. Hence, the proposed MRF model describes both intra-segmentation and inter-segmentation contextual information. Intra- neighborhood contains adjacent superpixels of a given segmentation. Inter- neighborhood contains intersecting superpixels from different segmentations. The MRF model is used to code contextual constraints in both intra- and inter- neighborhoods for more consistent labeling. We explore different data and neighborhood models for a more generalized contextual framework within the MRF formalization.

\begin{figure}[t]
\centering
\includegraphics[scale=0.70]{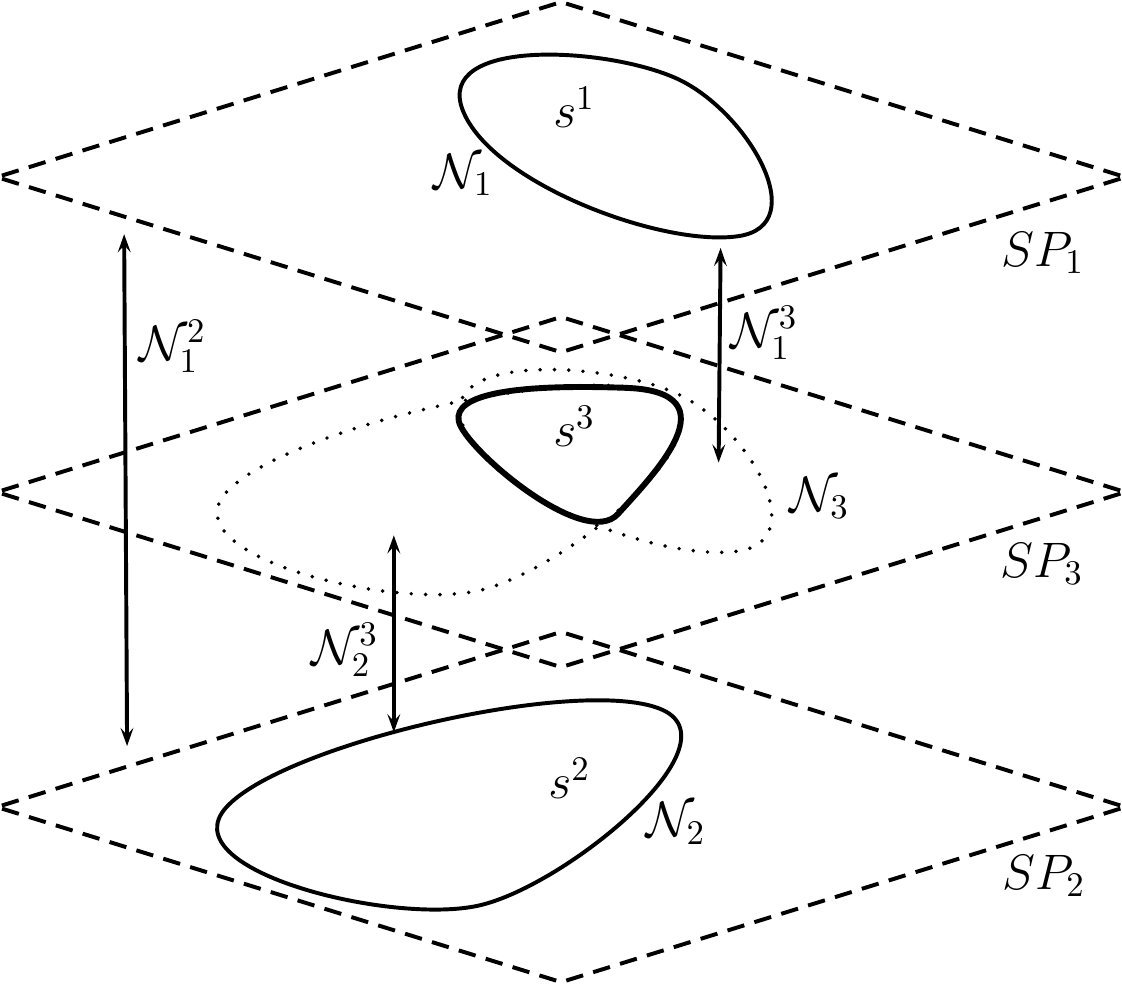}
\caption{Multi-hypothesis MRF model}
\label{mrf_figures}
\end{figure}

In the following, the generalized MRF model is described for two alternative segmentations; but it could be easily generalized to any number of alternatives. Let the set of segments/superpixels  be defined as $SP_m=\{s_i^m\}$ ($m=1,2$). We define a third segmentation $SP_3=\{s_k^3\}$ based on the intersection of the superpixels of the two alternatives (see Figure \ref{mrf_figures}):
 \begin{equation}
s_k^3 = s_i^1 \cap s_j^2 \neq \emptyset,  \:\:\:\: \forall \: s_i^1\in SP_1, \: s_j^2 \in SP_2
\end{equation}
For each segmentation, let $\mathcal{N}_m$ represent the contextual neighborhood that contains pairs of adjacent superpixels. In addition to these intra- neighborhoods, we define inter-segmentation context $\mathcal{N}_n^m (n,m=1,2,3)$ as follows:
\begin{equation}
(s_i^n,s_j^m) \in \mathcal{N}_n^m \:\:\mbox{ if }  \:\:s_i^n \cap s_j^m \neq \emptyset
\end{equation}
The image is parsed by assigning to each superpixel $s_i^m$ a class label $c_i^m$. Each segmentation produces an alternative parsing result, $\mathbf{c}^m =\{c_i^m\}$. We formulate the labeling problem as an MRF energy minimization over the whole set of superpixel labels $\mathbf{c}=\{\mathbf{c}^1,\mathbf{c}^2,\mathbf{c}^3\}$ as follows:
\begin{multline}
\label{gen_mrf}
\!\!\!\!\!J(\mathbf{c}) = \sum_{m=1}^3{\left(\sum_{s_i^m \in SP_m}\!\!\! D(s_i^m,c_i^m) + \lambda_m\!\!\!\!\!\sum_{(s_i^m,s_j^m) \in \mathcal{N}_m} \!\!\!\!\!\!E(c_i^m,c_j^m)\right)} \\ +  \sum_{(n,m)\in IC}{\lambda_n^m\!\!\!\!\!\sum_{(s_i^n,s_j^m) \in \mathcal{N}_n^m}\!\!\! E(c_i^n,c_j^m)}
\end{multline}
where $D$ and $E$ are appropriate data cost and smoothness cost terms, respectively, of related label assignments; $IC$ is the set of inter-segmentation neighborhoods ($IC=\{(1,2),(1,3),(2,3)\}$), and $\lambda_m, \lambda_n^m$ are smoothness constants for each corresponding context. In the next section we discuss the details of the MRF model, in particular how to define data and smoothness costs and how to select the smoothness constants.

\section{Image Parsing with Multi-hypothesis MRF Model}
\label{model}
In the MRF formulation given above, the data cost $D(s_i,c)$ represents the confidence with which a superpixel $s_i$ is assigned to a class $c$; the smoothness cost $E(c_i,c_j)$, on the other hand, is a measure of likelihood that two neighboring superpixels are assigned to distinct class labels $c_i,c_j$. Any parsing algorithm can be integrated into this MRF formulation, as long as it produces class-conditional likelihood scores that can be used to define the data costs of the model. In this paper we test three methods, SuperParsing, FCN and PSP, with optimized parameter settings.

SuperParsing is a data-driven, nonparametric parsing algorithm that tries to match the superpixels of a test image with the superpixels of a suitable subset of the training images, i.e. ``retrieval set'' (\cite{superparsing}).  A scene-level comparison is performed to find a good retrieval set that contains images similar to the tested image. For each test superpixel, a rich set of features are computed and matched with the superpixels from the retrieval set. The labels of these matching superpixels are used to compute class-conditional log-likelihood ratio scores, $L(s_i,c)$, for each class $c$. Then the data term is defined as $D(s_i,c) = w_i\sigma (L(s_{i},c))$, where  $w_i$ is the superpixel weight, and $\sigma(\cdot)$ is the sigmoid function. Here $w_i = |s_i| / \mu_s$, where $|s_i|$ is the  number of pixels in $s_i$, $\mu_s$ is the mean of $|s_i|$ (see \cite{superparsing} for details).

\begin{figure}[t]
\centering \subfloat[]{\epsfysize 2.3cm \epsfxsize 2.3cm
 \leavevmode
 \epsffile{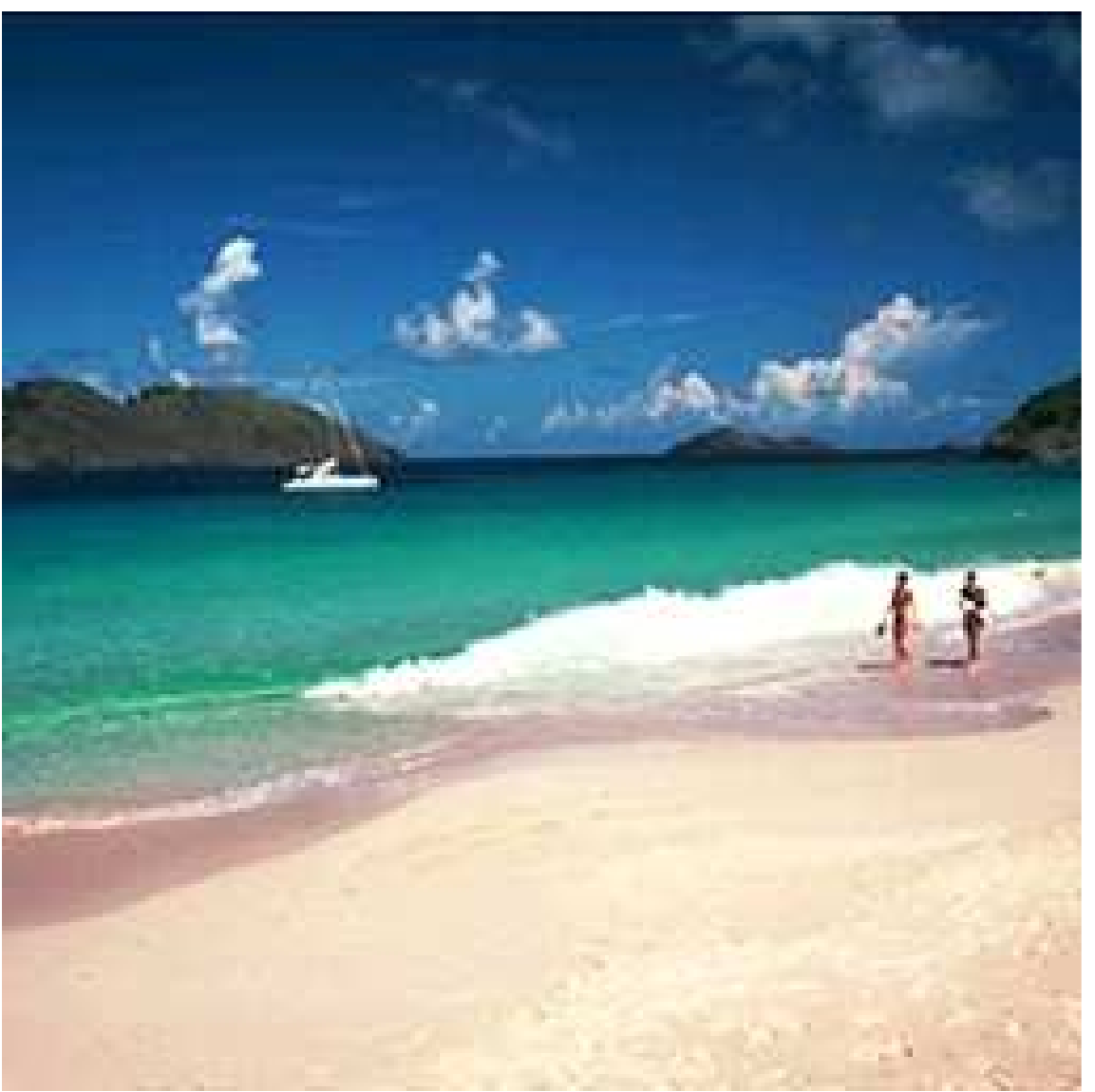}}
 \subfloat[]{\epsfysize 2.3cm \epsfxsize 2.3cm
 \leavevmode
 \epsffile{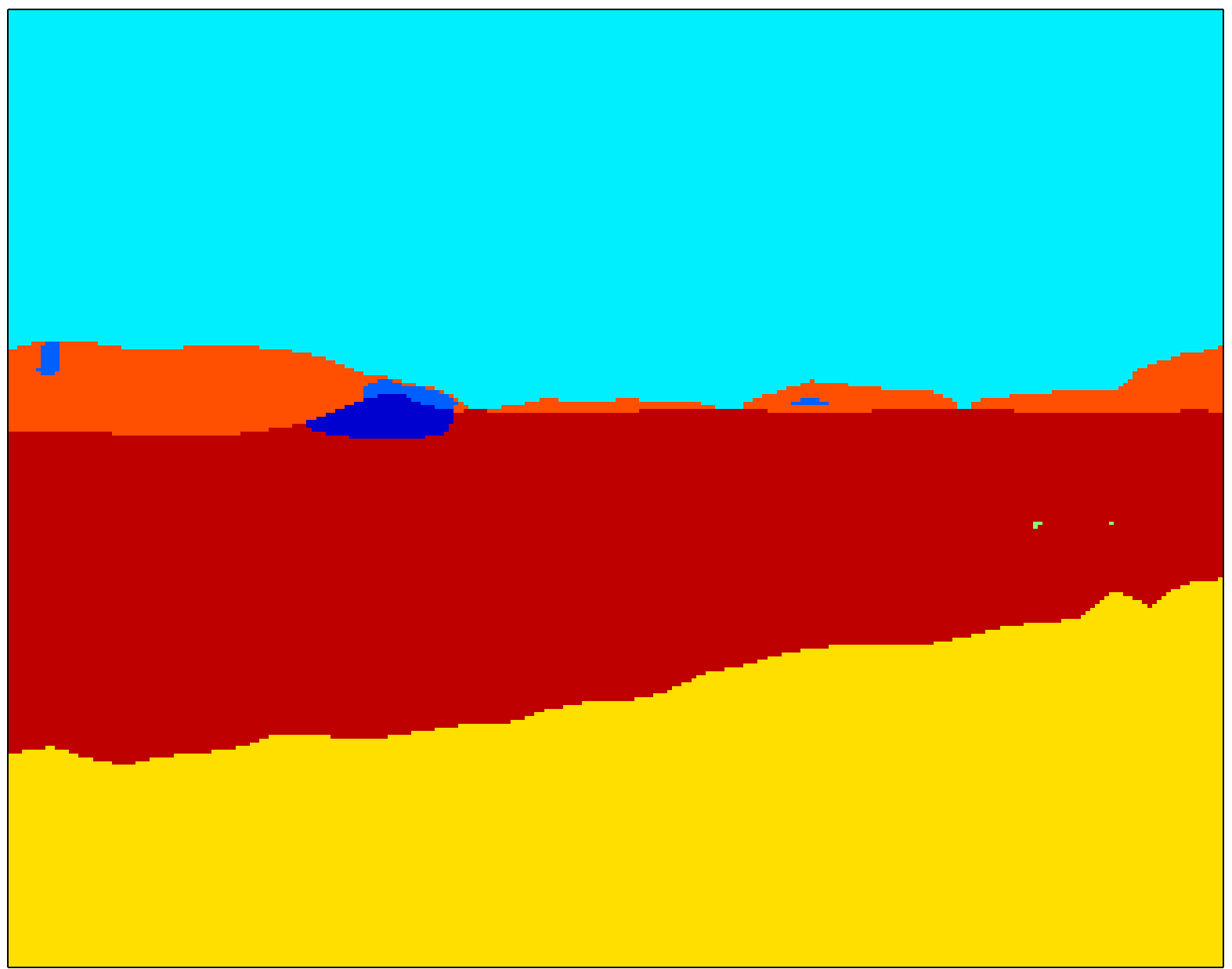} }
\subfloat{\epsfysize 2.3cm \epsfxsize 1.15cm
 \leavevmode
 \epsffile{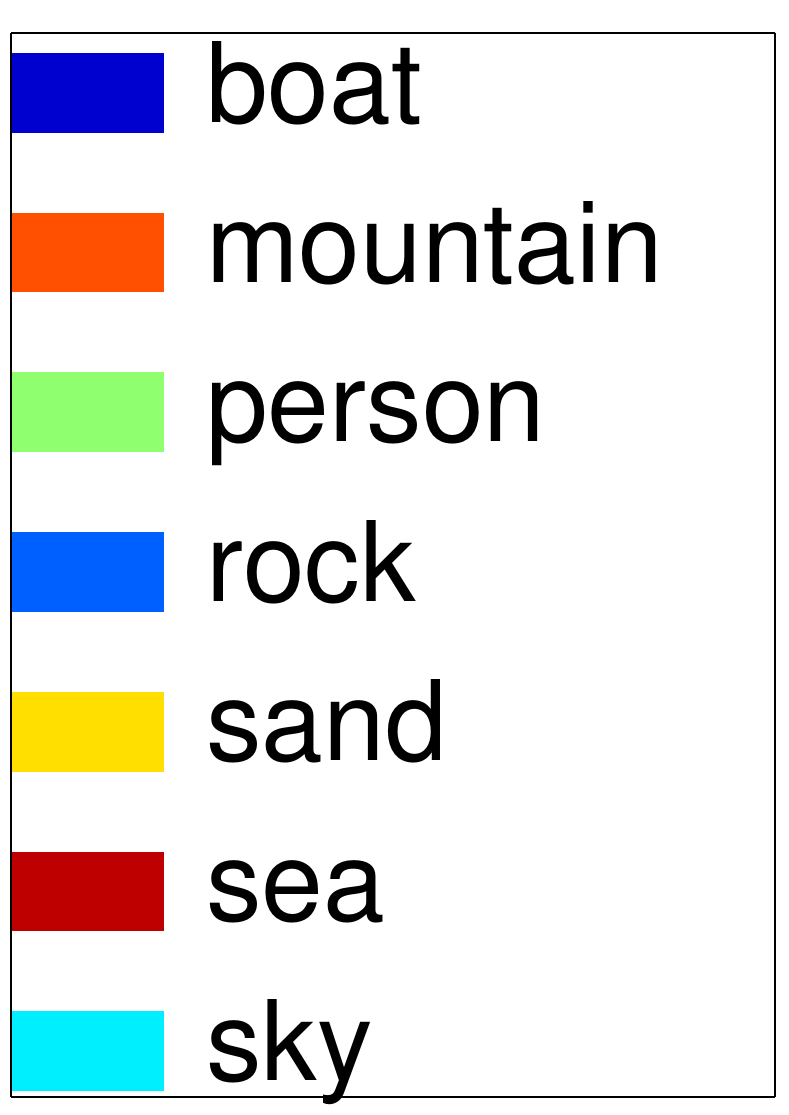}}\setcounter{subfigure}{2}
  \subfloat[]{\epsfysize 2.3cm \epsfxsize 2.3cm
 \leavevmode
 \epsffile{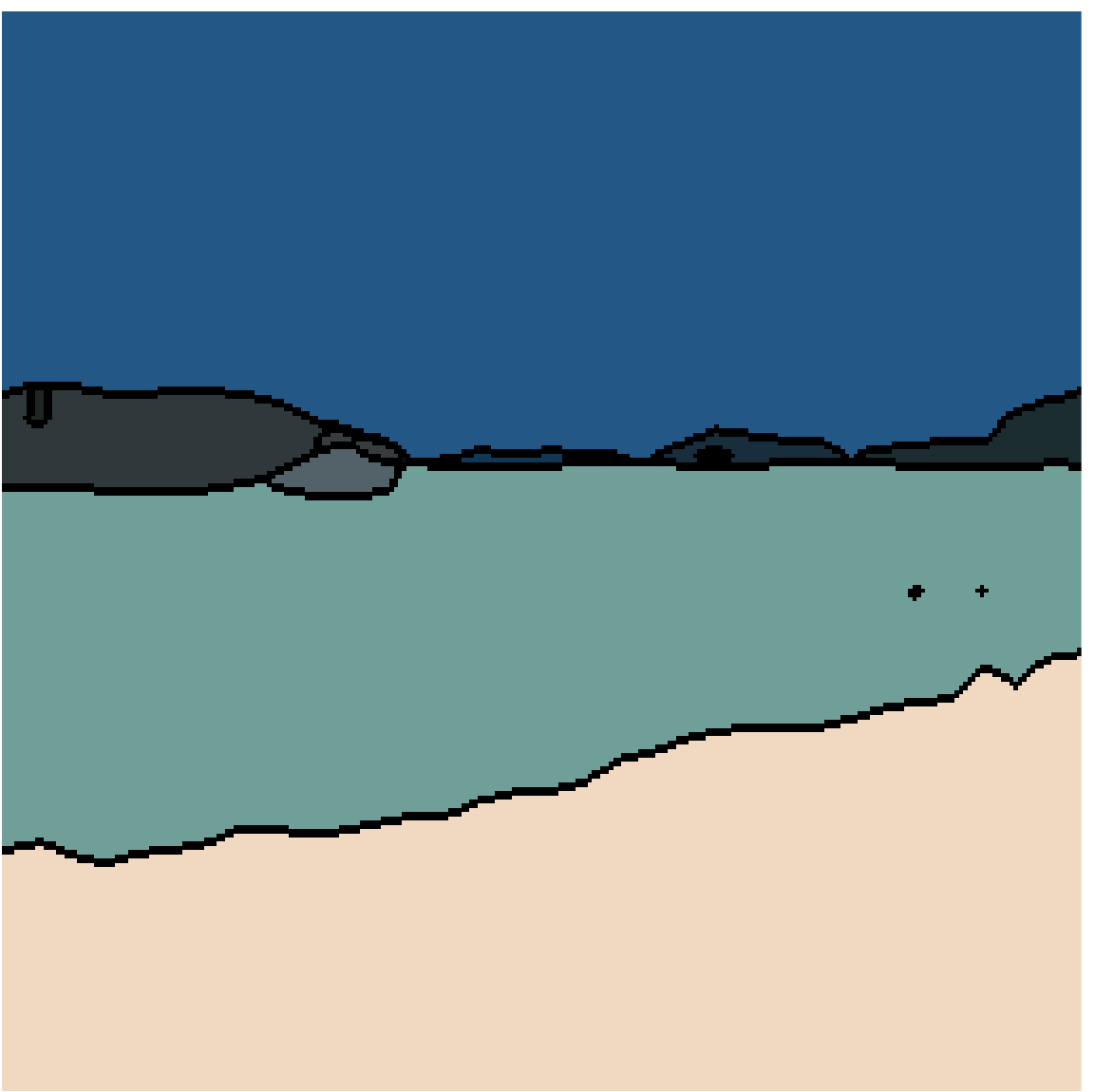}}
 \caption{FCN segmentation: (a) Original image; (b) FCN labels;  (c) FCN superpixels (colored by mean value of their pixels).}
 \label{cnn_seg}
\end{figure}

FCN/PSP architectures provide dense pixel-level class scores and labels at the output layer. These scores change smoothly between neighboring pixels, due to the interpolation at the deconvolution layers. We determine the connected components of the FCN/PSP outputs, where neighboring pixels with the same class label are assigned to the same component/segment (see Fig. \ref{cnn_seg}). Then the segment score is set equal to the mean score of its pixels. These scores are used to define the data terms of the MRF model, as described above. As seen in Fig. \ref{cnn_seg}(c), when the connected components are  very large, intra- and inter-segmentation neighborhoods will also be large and therefore will not capture the local context effectively. To overcome this, we intersect these connected components with superpixels of the image and test the use of intersecting regions in the MRF model, as well.

The smoothness costs for $SP_1$ and $SP_2$ are based on probabilities of label co-occurrence, as given in \cite{superparsing}:
\begin{equation}
E(c_i,c_j) \!=\! -\log[(P(c_i|c_j) + P(c_j | c_i))/2]\times\delta(c_i \neq c_j),
\end{equation}
where $P(c|c_j)$ is the conditional probability of one superpixel having label $c$ given that its neighbor has label $c_j$, estimated by counts from the training set. The delta function $\delta(c_i \neq c_j)$ is used to assign zero cost when $c_i=c_j$.

Since $SP_3$ is generated from the other two segmentations, its data and smoothness costs are defined as functions of the corresponding costs in $SP_1$ and $SP_2$. In order to fuse the complementary information coming from the two segmentations, we differentiate the set of classes in segment $SP_1$ and $SP_2$  and select from the union of those two sets for segment $SP_3$. In other words, there are two separate labels in $SP_3$, $c^{(1)}$ and $c^{(2)}$,  which correspond to the semantic class $c$ coming from $SP_1$ and $SP_2$, respectively (Note that, $c_i^m$ represents the class label assigned to superpixel $s_i^m$, while $c^{(m)}$ represents any given class of segmentation $SP_m$). Then the data cost for $s_k^3 \in SP_3$ is given by ($m=1,2$):
 \begin{equation}
D(s_k^3,c^{(m)}) = f_m(D(s_i^1,c), D(s_j^2,c)),
\end{equation}
where $(s_i^1,s_k^3) \in \mathcal{N}_1^3$ and $(s_j^2,s_k^3) \in \mathcal{N}_2^3$.
Likewise, the smoothness costs in inter- neighborhoods $\mathcal{N}_n^m$ are based on the costs in $\mathcal{N}_1$ and $\mathcal{N}_2$ as follows:
 \begin{equation}
E(c_i^{(1)},c_j^{(2)}) = g(E(c_i^{(1)},c_j^{(1)}), E(c_i^{(2)},c_j^{(2)}))
\end{equation}
The specifics of $f_m$ and $g$ will be discussed in the next section.

The smoothness constants $\lambda_m, \lambda_n^m$ control the level of contextual dependency in different neighborhoods of the MRF model in Eq. \eqref{gen_mrf}. These values are determined from the training set by a leave-one-out strategy: each training image is removed from the training set, and  then parsed by the proposed algorithm to obtain its labeling accuracy under different parameter settings. Then, the set of parameters that maximize the mean pixel accuracy in the training set is chosen.

The MRF energy function is minimized by the $\alpha$-expansion method of \cite{Min_cut}. The outcome leads to three alternative labelings $\mathbf{c}=\{\mathbf{c}^1,\mathbf{c}^2,\mathbf{c}^3\}$. The set of labels $\mathbf{c}^3$ of the segmentation $SP_3$ is selected as the final labeling of the image.

\section{Simulations and Discussions}
\label{benzetim}
\subsection{Implementation Details}
The smoothness constants of the MRF model are selected from the set $\{l\lambda | \:\: l\in\{0,1,2\}; \:\: 5\leq\lambda\leq 25, \lambda\in\mathbb{Z}\}$. The function $g$ is set as $g(x,y)=0.5x+0.5y$. For data costs of $SP_3$, we test two alternatives :
\begin{itemize}
 \item {\bf DC1}: ($m=1,2$)
 \begin{equation}
 D(s_k^3,c^{(m)}) = \beta_1\frac{w_k}{w_i^1}D(s_i^1,c) + \beta_2\frac{w_k}{w_j^2}D(s_j^2,c)
 \end{equation}
 \item {\bf DC2}:
 \begin{equation}
 \begin{aligned}
 D(s_k^3,c^{(1)}) &= \frac{w_k}{w_i^1}D(s_i^1,c) \\
 D(s_k^3,c^{(2)}) &= \frac{w_k}{w_j^2}D(s_j^2,c)
\end{aligned}
\end{equation}
\end{itemize}
where
$w_k=0.5|s_k^3|(1/|s_i^1|+1/|s_j^2|)$ and $\beta_1+\beta_2=1$.

The first model uses a weighted average of two data costs from $SP_1$ and $SP_2$ for superpixels of $SP_3$ and does not differentiate labels $c^{(1)}$ and $c^{(2)}$. The second model assigns two different data costs to the labels $c^{(1)}$ and $c^{(2)}$ of the same class $c$ under two different hypotheses of $SP_1$ and $SP_2$. Therefore the second model enables the algorithm to choose between the two hypotheses, depending on the contextual information.
The superpixel weight $w_k$ is set proportional to the relative size of $s_k^3$ with respect to $s_i^1$ and $s_j^2$; if the intersection of the two superpixels is  small, then the data cost for $s_k^3$ is assigned lower weight since it is deemed unreliable for labeling $s_k^3$. The data weights $\beta_m$ represent the confidence in the segmentation result of the two hypotheses; hence the better algorithm is assigned a higher weight.

SuperParsing uses graph-based superpixel segmentation (\cite{graph}). In this algorithm, parameter $K$ controls superpixel color consistency and $S$ determines the smallest superpixel size. $R$ is the number of nearest neighbor matches for each superpixel feature. Several superpixel features, including shape (e.g. superpixel area), location (e.g. superpixel mask), texture (e.g. SIFT), color (e.g. color histogram) descriptors, are used to find matching superpixels from the retrieval set. There are six SIFT features that are defined over different subregions of the superpixel. These SIFT features are encoded  using LLC (Locality-constrained Linear Coding) (\cite{LLC}) or KCB (Kernel Codebook Encoding) (\cite{KCB}) algorithms.

In addition, we use CNN features both for global matching to determine the retrieval set and for local superpixel matching. These are learned features extracted from the last layer of the network before the final classification layer. CNN features are  extracted from the trained networks of VGG-F (\cite{Chatfield14}) and AlexNet (\cite{NIPS2012_4824}), which are trained on ILSVRC ImageNet dataset (\cite{imagenet}). For each superpixel, its CNN feature is obtained by setting the whole image to zero except for the superpixel region and computing the network output.

In the following section, simulation results are provided for the following three alternative segmentations:
\begin{itemize}
\item {\bf Seg1}: SuperParsing (K=200, S=100, R=30). VGG-F/AlexNet features are used both for global matching and for superpixel matching.


\item {\bf Seg2}: FCN-8s segmentation (\cite{fcnn}) is used to define superpixels and data costs, as described in the previous section.
\item {\bf Seg3}: PSP segmentation, no cropping, single scale evaluation (\cite{psp}), usage similar to FCN-8s.
\end{itemize}


\subsection{Results and Comparisons}
The proposed models are evaluated on two well-known datasets, namely SIFT Flow (\cite{siftflow})  and 19-class subset of LabelMe (\cite{labelmesub}). In experiments overall pixel-level classification accuracy (i.e. correctly classified pixel percentage) and average per-class accuracies are compared.

SIFT Flow dataset contains 2,688 images and 33 labels. This dataset includes outdoor scenery such as mountain view, streets, etc. There are objects from 33 different semantic classes, such as {\it sky}, {\it sea}, {\it tree}, {\it building}, {\it cars}, at various densities. There are also 3 geometric classes, which are not considered in our simulations. Dataset is separated into 2 subsets; 2,488 training images and 200 test images. The retrieval set size is set at 200 images, as in original SuperParsing. LLC encoding is used for SIFT features.

Table \ref{table:accuracy} reports the pixel-level and average per-class labeling accuracies of DC1 and DC2 models with optimized parameter settings.
We also provide Seg1, Seg2 and Seg3 results with MRF smoothing, but without using the inter- neighborhoods (i.e. $\lambda_n^m =0, \:\: \forall (n,m)\in IC$). For comparison, results are provided for original SuperParsing, original FCN-8s, PSP and for some other recent superpixel-based and CNN-based methods evaluated on SIFT Flow dataset.

\begin{table}[t]
\centering
\small
\caption{Per-pixel and average per-class labeling accuracies for SIFT Flow dataset}
\vspace{-0.1cm}
\begin{tabular}{|l||c||c|}
\hline
                   &  \multicolumn{2}{c|}{Percentage Accuracy}              \\ \hline
{\it Method}            &   Per-pixel (\%)    &  Per-class (\%)        \\ \hline \hline
DC1 (Seg1,Seg2)          & 86.8  &  50.8              \\
DC2 (Seg1,Seg2)          & 87.2  &  50.4              \\
DC1 (Seg2,Seg3)          & {\bf 88.7}  &  55.2              \\
DC2 (Seg2,Seg3)          & 88.4  &  54.2              \\\hline
Seg1+MRF                 & 80.9  &  37.2              \\
Seg2+MRF                 & 86.0  &  51.6                 \\
Seg3+MRF                 & 87.8  &  50.0                 \\ \hline
SuperParsing             & 76.2  &  29.1              \\
FCN-8s (Seg2)            & 85.9  &  53.9              \\
PSP    (Seg3)            & 87.7  &  51.7              \\ \hline
\cite{Ates2017}          & 80.6  &  31.8                \\
\cite{fischer}           & 81.7  &  50.1                \\
\cite{nguyen_cnn}        & 83.3  &  49.4                \\
\cite{parsenet}          & 86.8  &  52.0               \\
\cite{stacked}           & 86.4  &  49.4                \\
\cite{dag_rcnn}          & 85.3  &  {\bf 55.7}              \\ \hline
\end{tabular}
\vspace{-0.3cm}
\label{table:accuracy}
\end{table}


Table \ref{table:accuracy} includes results for two alternatives, i.e. (Seg1,Seg2) and (Seg2,Seg3). (Seg1,Seg3) combination is inferior to (Seg2, Seg3) and hence not reported. For (Seg1,Seg2), DC2 improves the pixel accuracy of FCN-8s by 1.3\% through the use of proposed multi-hypothesis MRF model. DC1 accuracy is 0.4\% lower than DC2, showing that it is more effective to keep data costs separate when fusing the complementary information of the two hypotheses. Note that, without the inter- neighborhoods, Seg2+MRF could only achieve 0.1\% improvement over FCN-8s through MRF smoothing. This shows that our proposed MRF framework successfully fuses the segmentation decisions of alternative methods. In addition, when compared to our previous results in \cite{Ates2017}, parsing accuracy is substantially better, due to use of CNN features and FCN segmentation.

For (Seg2,Seg3), DC1 is the better model and improves the pixel accuracy of PSP by 1.0\%. DC2 accuracy is 0.3\% lower than DC1 this time. When combining FCN and PSP results, the connected components are intersected with superpixels of Seg1 to provide finer segmentation, as explained in Section \ref{model}. This improves both pixel accuracy and average per-class accuracy of MRF fusion. The accuracy of Seg3+MRF is much lower than our results, once again showing the importance of inter- neighborhoods in the model. As seen from Table \ref{table:accuracy}, DC1(Seg2,Seg3) outperforms  other recent superpixel-based and CNN-based segmentation methods in terms of per-pixel labeling accuracy,

The average per-class accuracy of DC1(Seg2,Seg3) is 1.3\% better than FCN-8s and 3.5\% better than PSP. However, for (Seg1,Seg2), the average per-class accuracies of both DC2 and DC1 are lower than that of FCN-8s, both due to the spatial smoothing of labels by MRF optimization and also because mean class accuracy of Seg1 is significantly lower and therefore not helping to boost the overall performance.

Note that using MRF smoothing alone on both FCN and PSP outputs reduces mean class accuracies with marginal improvement on per-pixel accuracies. This is because MRF optimization favors dominant classes covering large areas (such as {\it sky}, {\it building}) and smooths out rare classes with smaller areas (such as {\it car}, {\it window}). As a result per-class accuracies of dominant classes are slightly increased at the expense of significant drop in per-class accuracies of rare classes. On the other hand, our MRF framework is capable of increasing both per-pixel and per-class accuracies, when smoothness constants in the model are carefully selected and when both hypotheses have comparable performance.
Therefore we believe that per-class accuracy of (Seg1,Seg2) could also be improved within the proposed MRF framework, if Seg1 is assigned to a better performing superpixel-based parsing algorithm (such as \cite{fischer}).

\begin{table*}[t]
\centering
\small
\caption{Per-class labeling accuracies for selected classes of SIFT Flow dataset}
\vspace{-0.1cm}
\begin{tabular}{|l||c|c|c|c|c|c|}
\hline
                   &  \multicolumn{6}{c|}{Per-class Accuracy (\%)}              \\ \hline
{\it Class}     & Seg1+MRF & FCN-8s &  Seg2+MRF     & PSP & DC2(Seg1,Seg2) & DC1(Seg2,Seg3)  \\ \hline \hline
{\it sky}       & 93.9  &  97.3 & 97.4 & 97.5 & 97.5 & 98.0           \\
{\it building}  & 90.4  &  91.8 & 92.9 & 94.6 & 94.0 & 95.2       \\ \hline
{\it car}       & 66.3  &  86.1 & 84.7 & 89.7 & 84.4 & 91.0               \\
{\it window}    & 28.4  &  50.8 & 35.3 & 18.1 & 48.4 & 33.3            \\
{\it person}    &  3.1  &  25.8 & 24.7 & 34.9 & 26.1 & 39.4     \\ \hline
\end{tabular}
\vspace{-0.3cm}
\label{table:acc_class}
\end{table*}

Table \ref{table:acc_class} lists per-class accuracies of the tested algorithms for some dominant ({\it sky}, {\it building}) and rare ({\it car}, {\it window}, {\it person}) classes. As explained above, MRF smoothing improves the parsing performance of dominant classes that cover large regions in tested images; however many segments from rare classes are also mistakenly assigned to labels of their dominant neighbors (e.g. {\it window} vs. {\it building}), which decreases the accuracies of these rare classes. While (Seg1,Seg2) also suffers from this problem due to the poor performance of Seg1, rare class accuracies of (Seg2,Seg3) are generally better than those of both FCN and PSP. Therefore our MRF framework is effective in combining the best of both segmentations without over-smoothing the label assignments.

\begin{table}[t]
\centering
\small
\caption{Per-pixel and average per-class labeling accuracies for 19-class LabelMe dataset}
\vspace{-0.1cm}
\begin{tabular}{|l||c||c|}
\hline
                   &  \multicolumn{2}{c|}{Percentage Accuracy}              \\ \hline
{\it Method}            &   Per-pixel (\%)    &  Per-class (\%)        \\ \hline \hline
DC1 (Seg1,Seg2)                  &  {\bf 90.4} & 75.5               \\
DC2 (Seg1,Seg2)                  &  89.9 & {\bf 76.1}           \\ \hline
Seg1+MRF                         &  86.7 & 67.3               \\
Seg2+MRF                         &  89.0 & 72.5               \\ \hline
FCN-8s                           &  86.9 & 70.0              \\ \hline
\cite{Ates2017}                  &  83.8 & 55.6          \\
\cite{Myeong2013}                &  81.8 & 54.4          \\
\cite{adapt_super}               &  82.7 & 55.1          \\
\cite{nguyen_cnn}                &  85.5 & 63.2          \\ \hline
\end{tabular}
\vspace{-0.3cm}
\label{table:accuracyLMS}
\end{table}

\begin{figure*}[t]
\centering \subfloat{\epsfysize 2.5cm \epsfxsize 2.5cm
 \leavevmode
 \epsffile{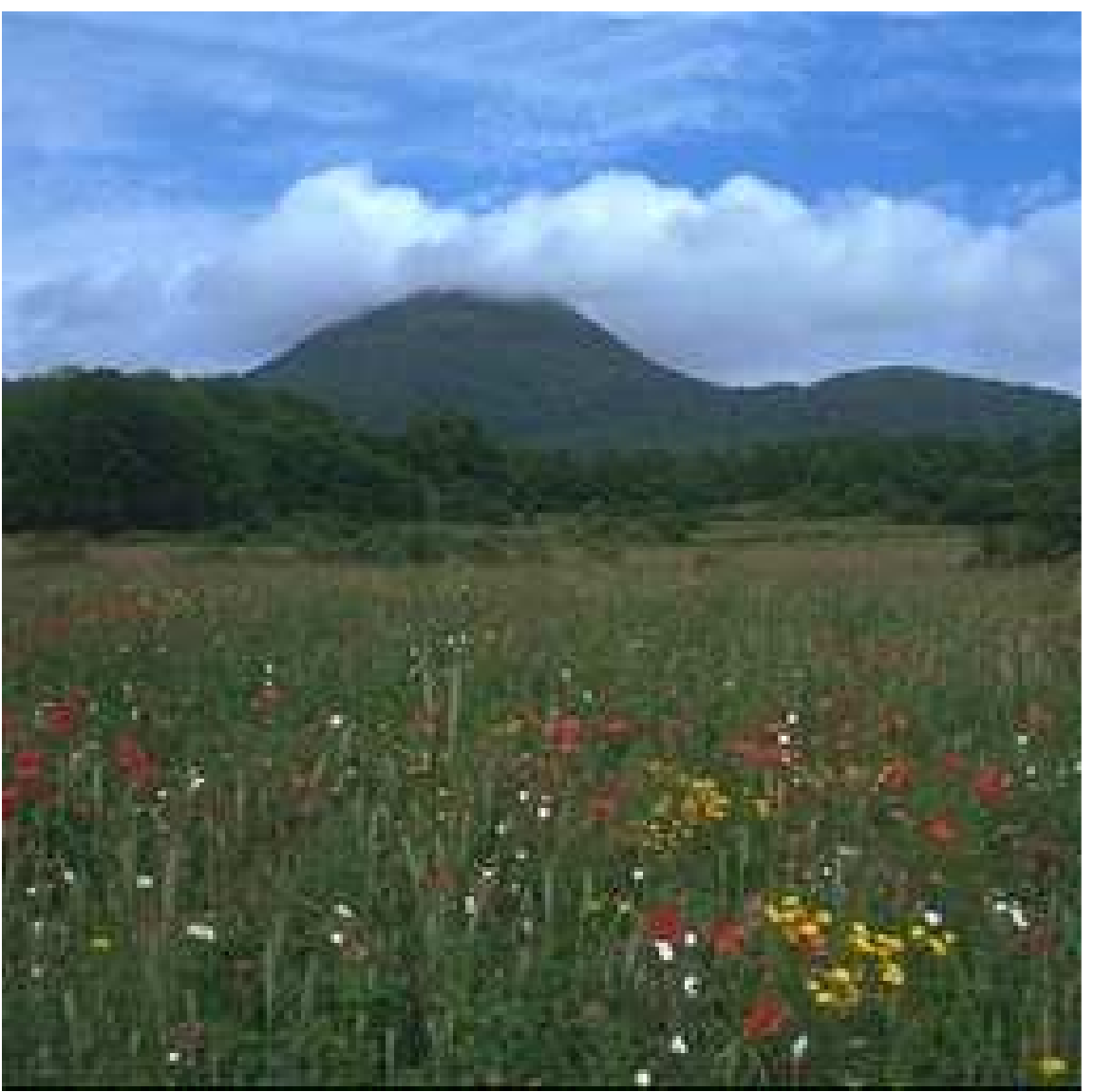}}
 \subfloat{\epsfysize 2.5cm \epsfxsize 2.5cm
 \leavevmode
 \epsffile{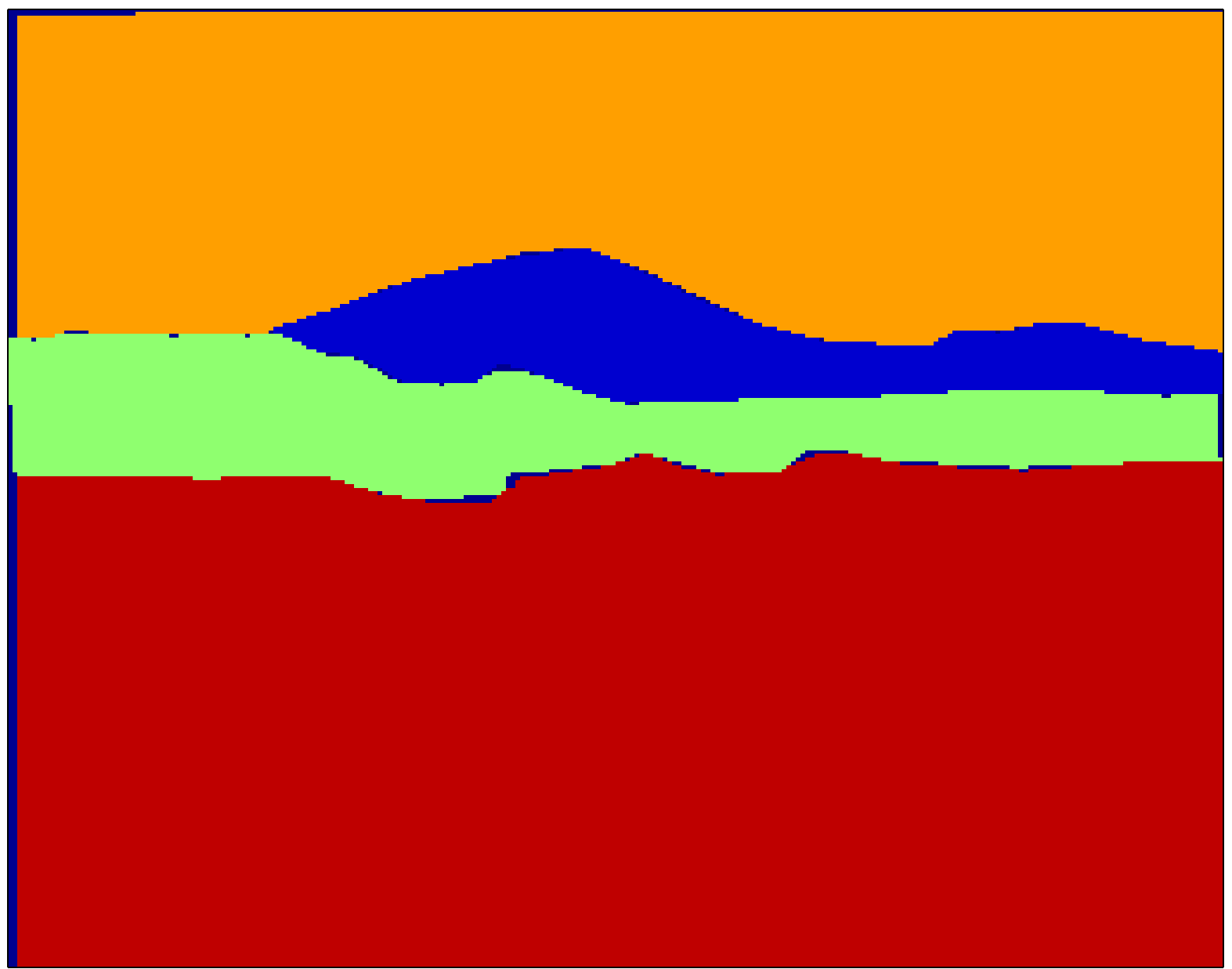}}
 \subfloat{\epsfysize 2.5cm \epsfxsize 2.5cm
 \leavevmode
 \epsffile{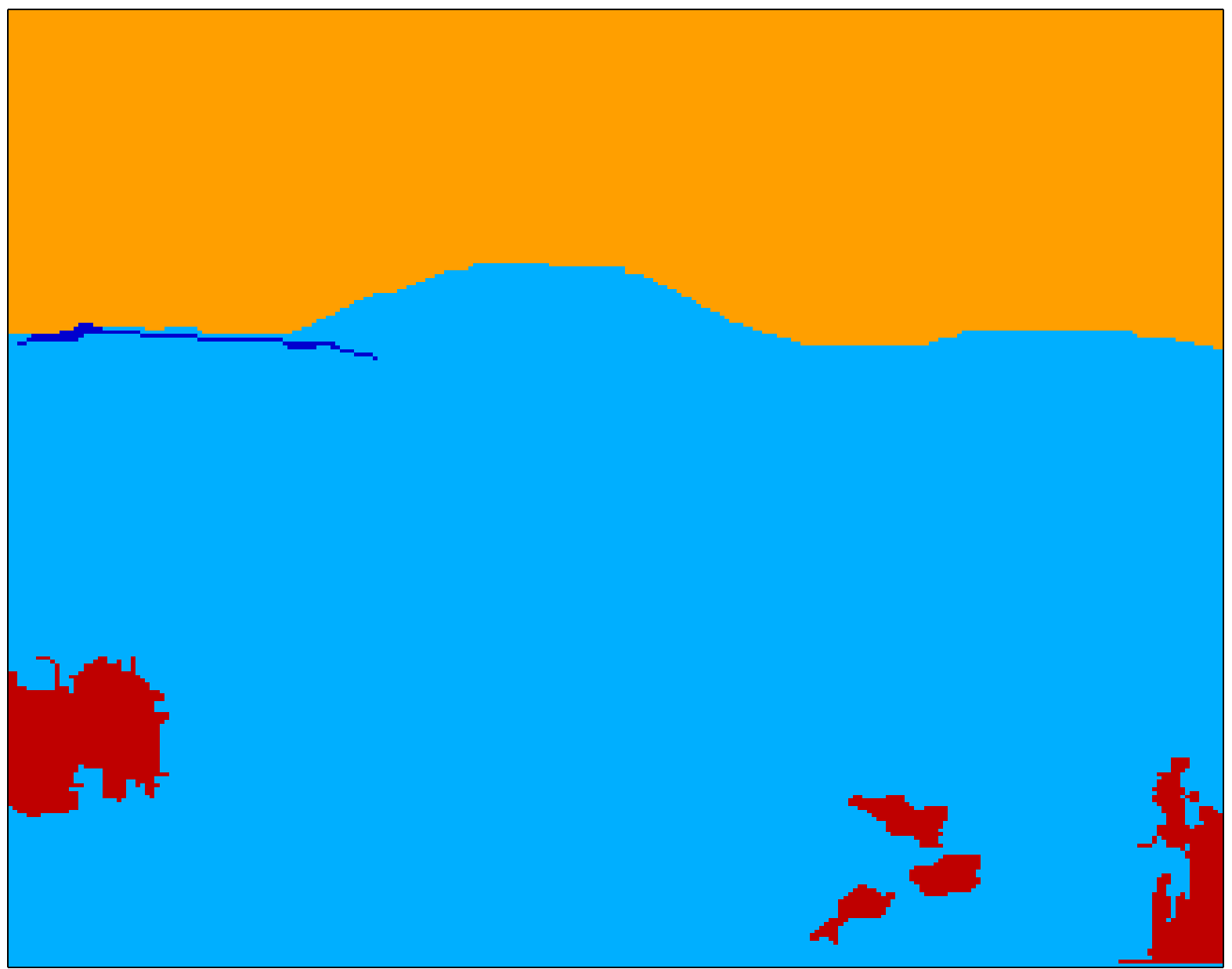}}
\subfloat{\epsfysize 2.5cm \epsfxsize 2.5cm
 \leavevmode
 \epsffile{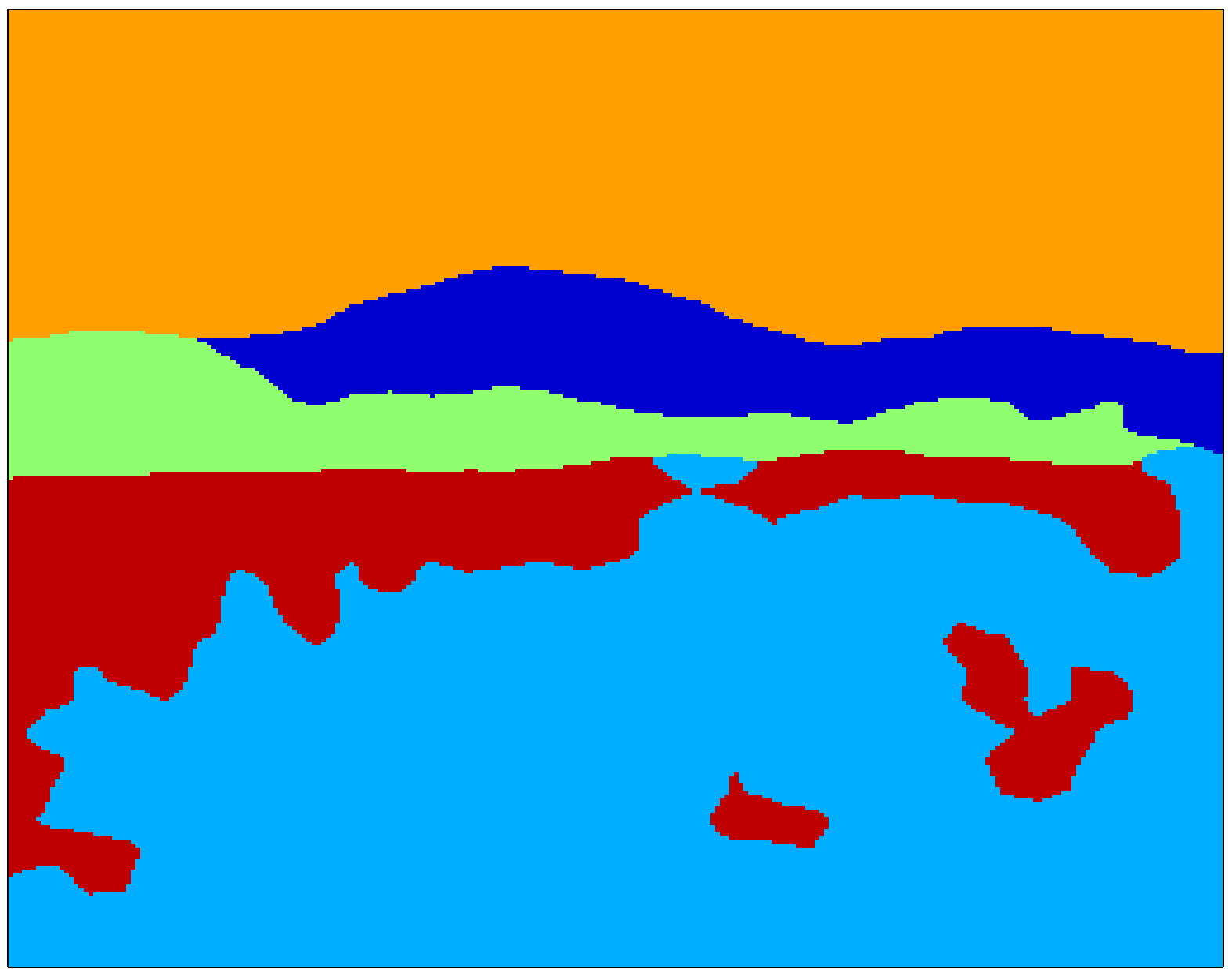}}
  \subfloat{\epsfysize 2.5cm \epsfxsize 2.5cm
 \leavevmode
 \epsffile{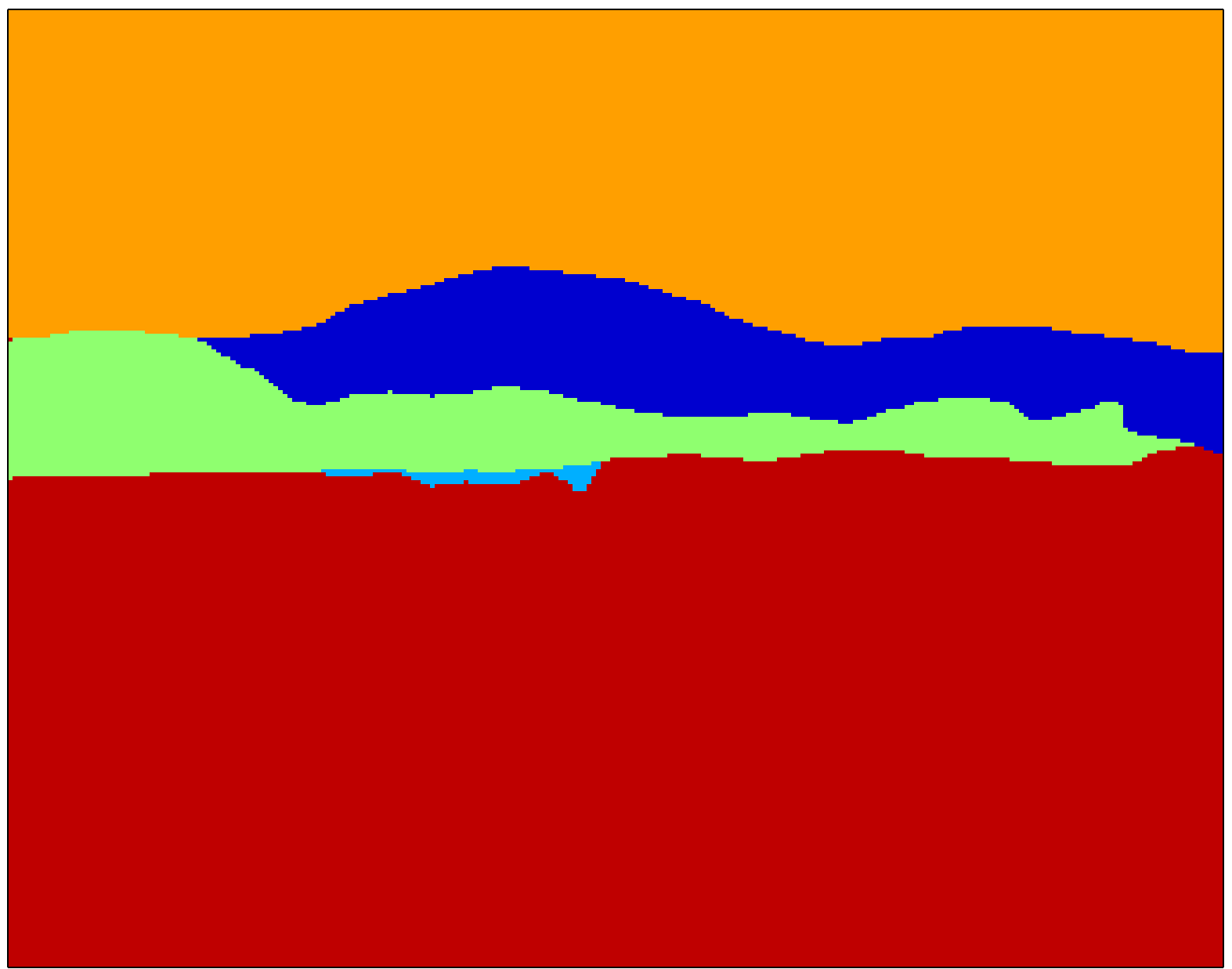}}
  \subfloat{\epsfysize 2.5cm \epsfxsize 1.25cm
 \leavevmode
 \epsffile{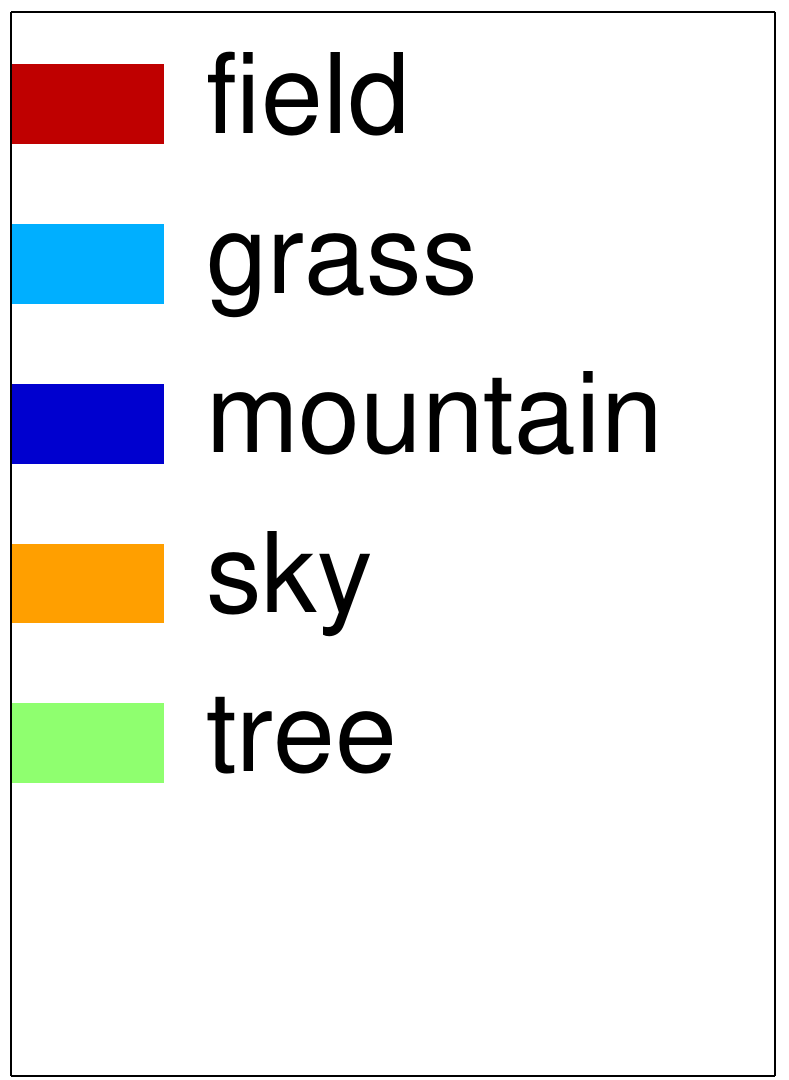}}

\centering \subfloat{\epsfysize 2.5cm \epsfxsize 2.5cm
 \leavevmode
 \epsffile{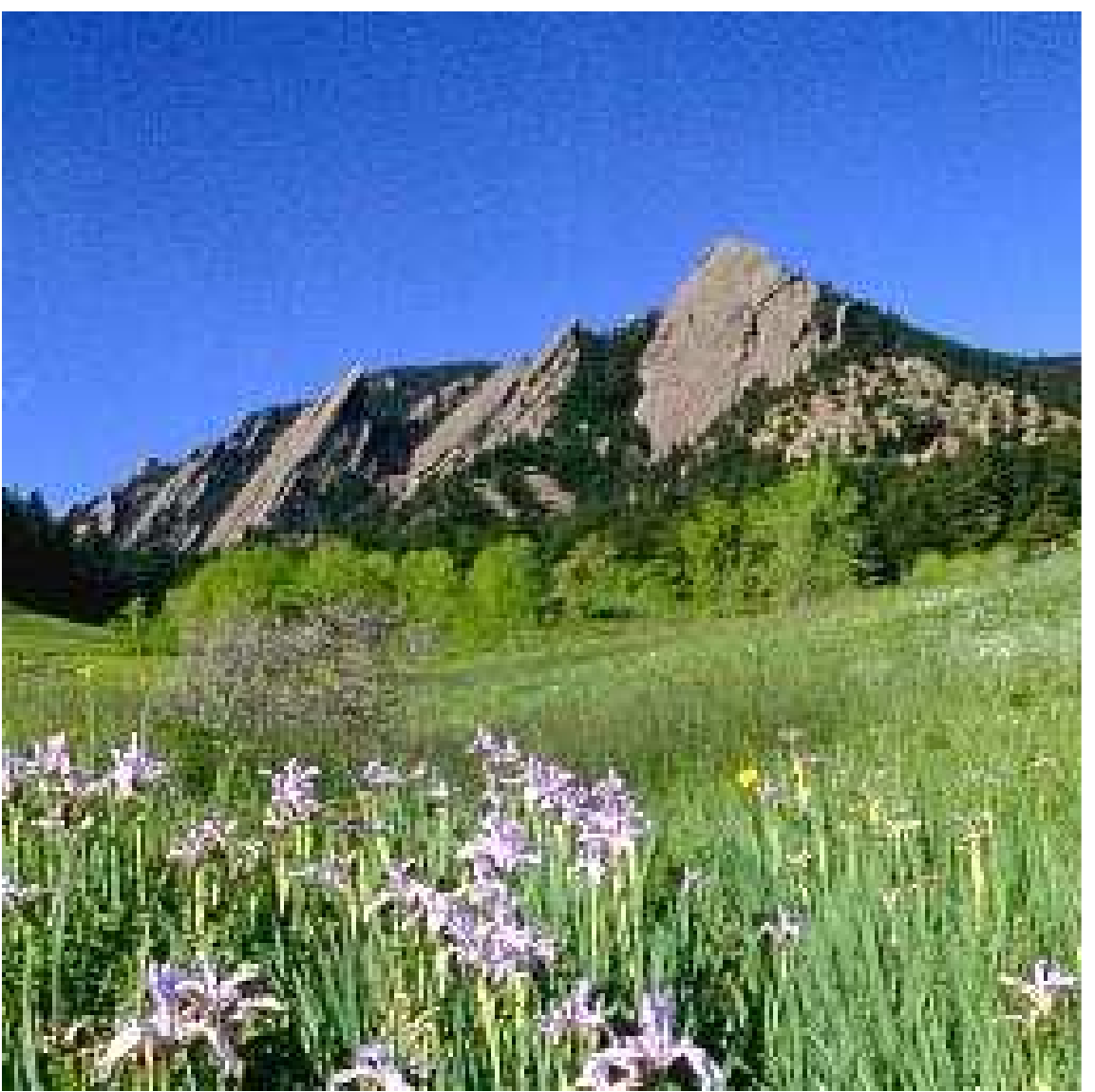}}
 \subfloat{\epsfysize 2.5cm \epsfxsize 2.5cm
 \leavevmode
 \epsffile{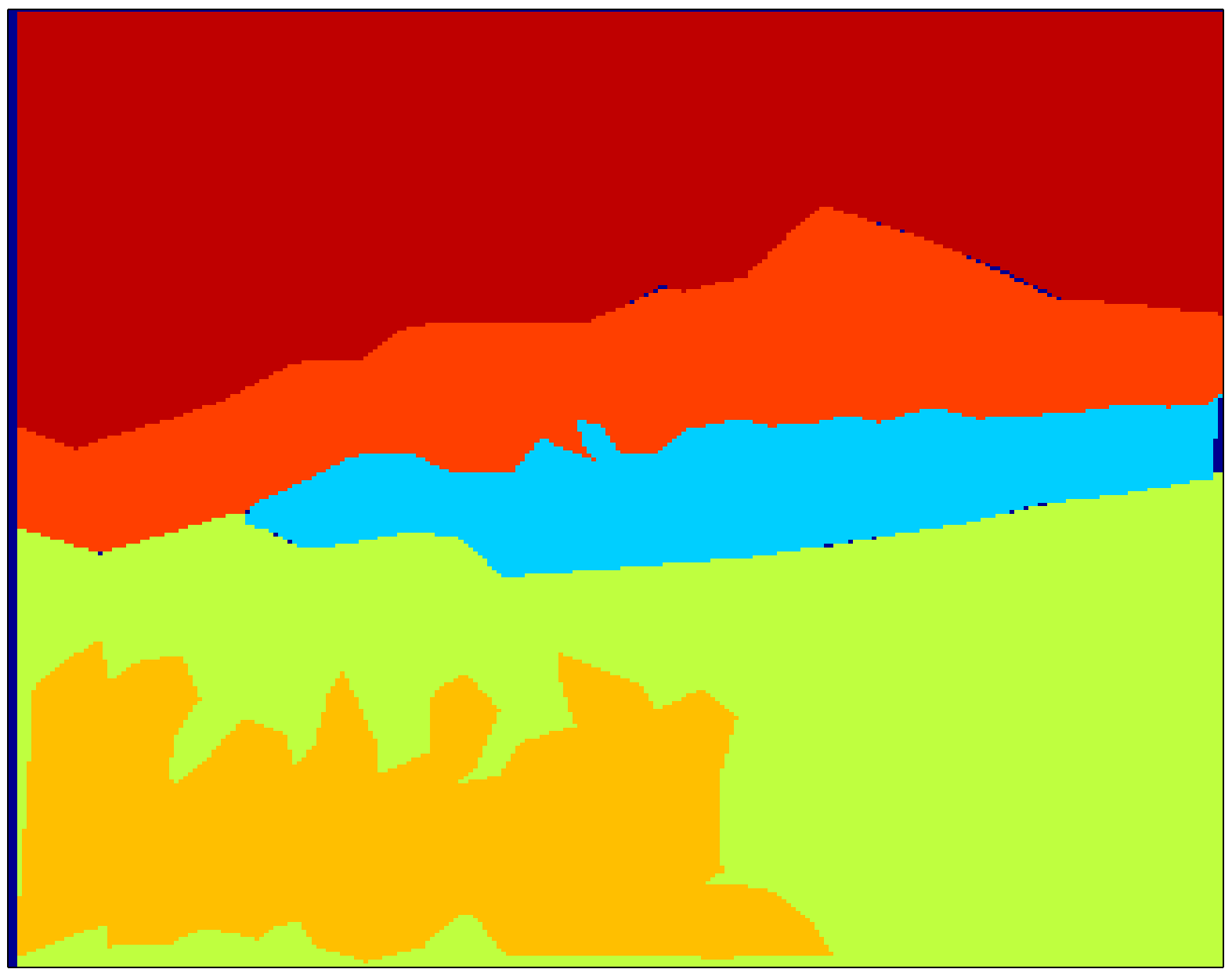}}
 \subfloat{\epsfysize 2.5cm \epsfxsize 2.5cm
 \leavevmode
 \epsffile{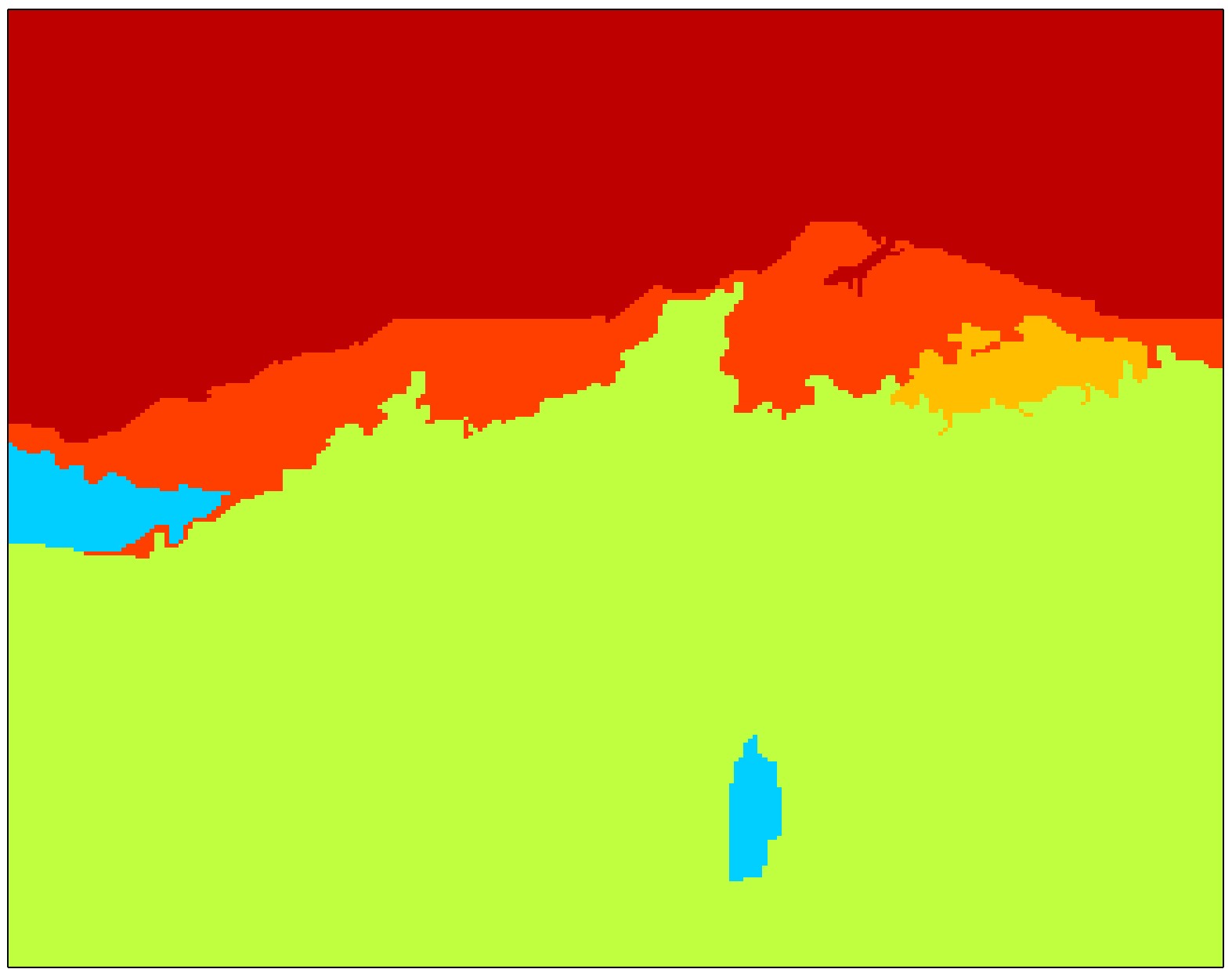}}
\subfloat{\epsfysize 2.5cm \epsfxsize 2.5cm
 \leavevmode
 \epsffile{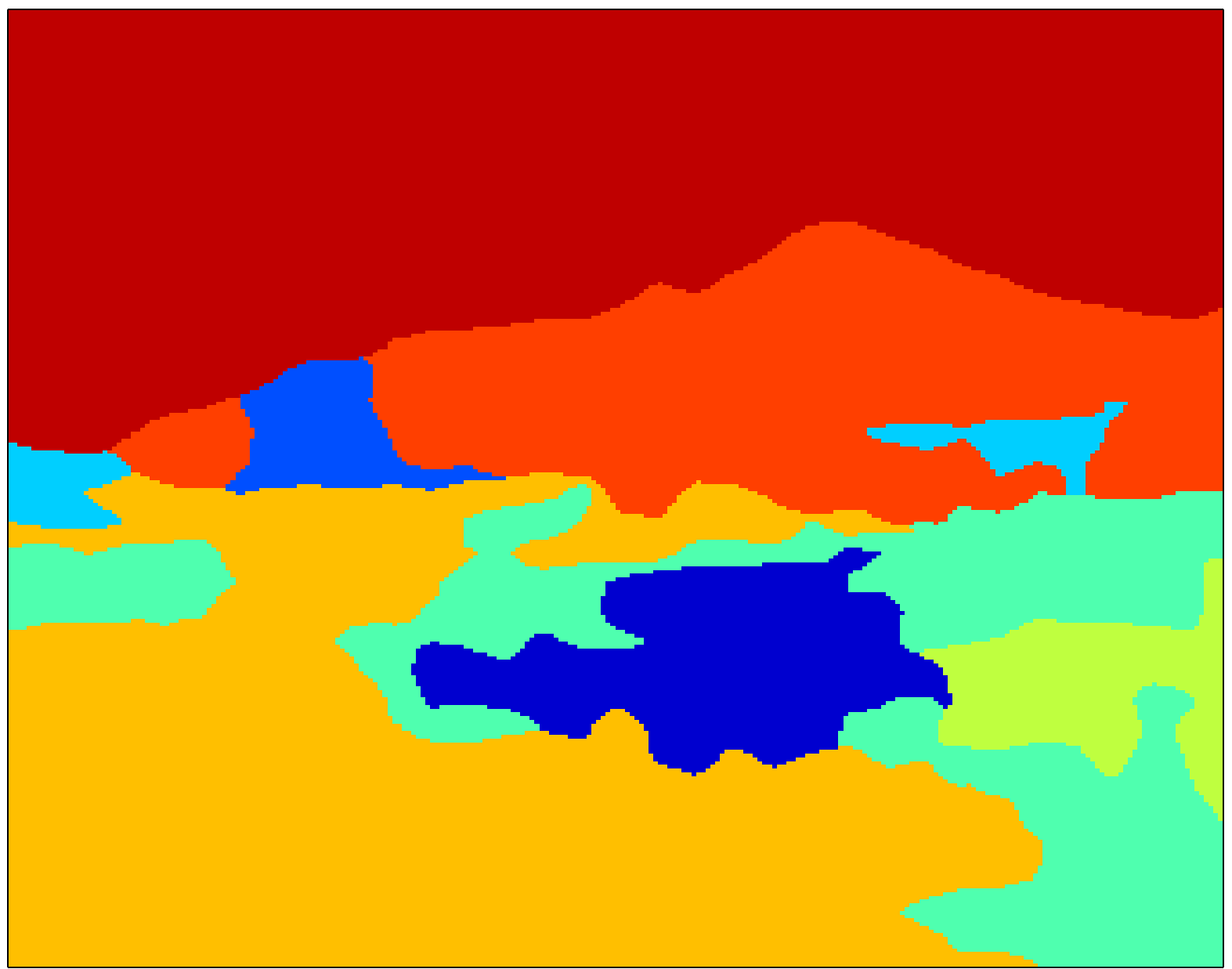}}
  \subfloat{\epsfysize 2.5cm \epsfxsize 2.5cm
 \leavevmode
 \epsffile{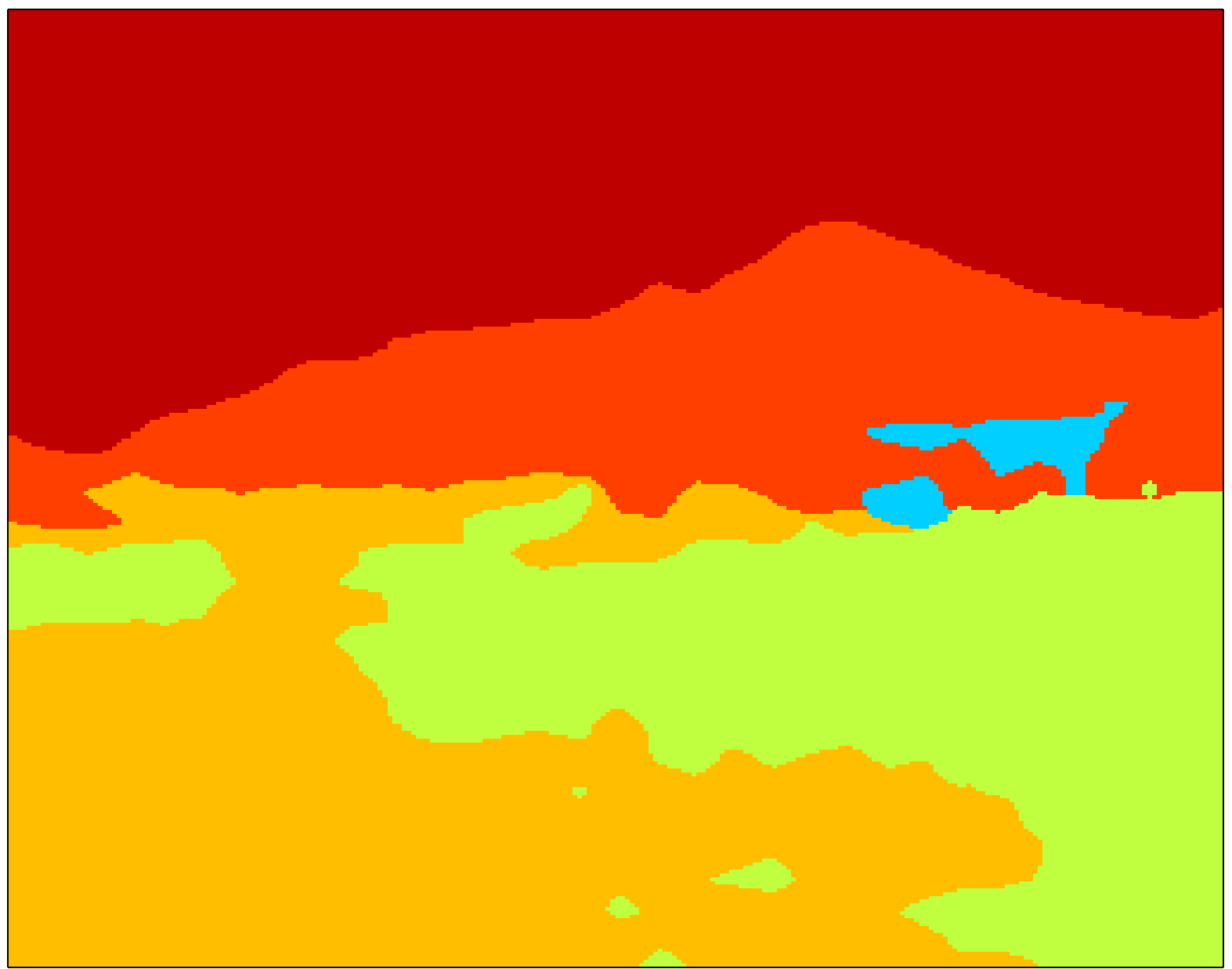}}
  \subfloat{\epsfysize 2.5cm \epsfxsize 1.25cm
 \leavevmode
 \epsffile{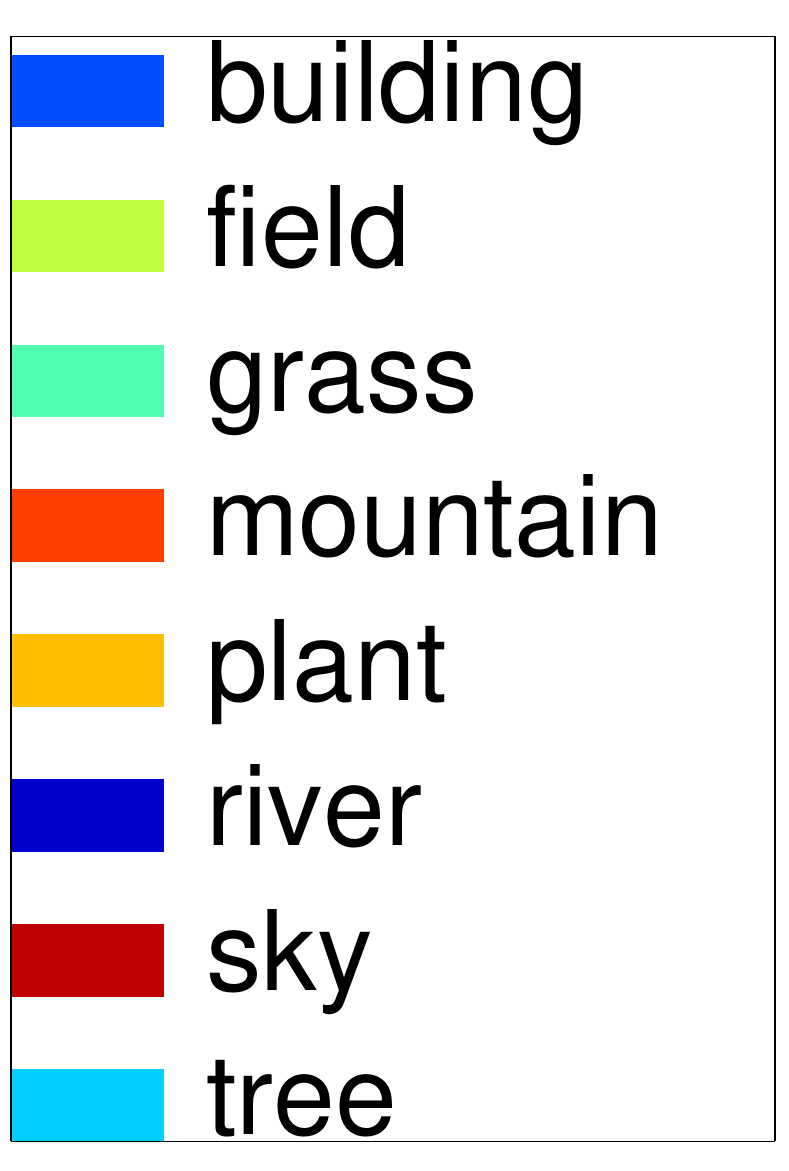}}

\centering \subfloat{\epsfysize 2.5cm \epsfxsize 2.5cm
 \leavevmode
 \epsffile{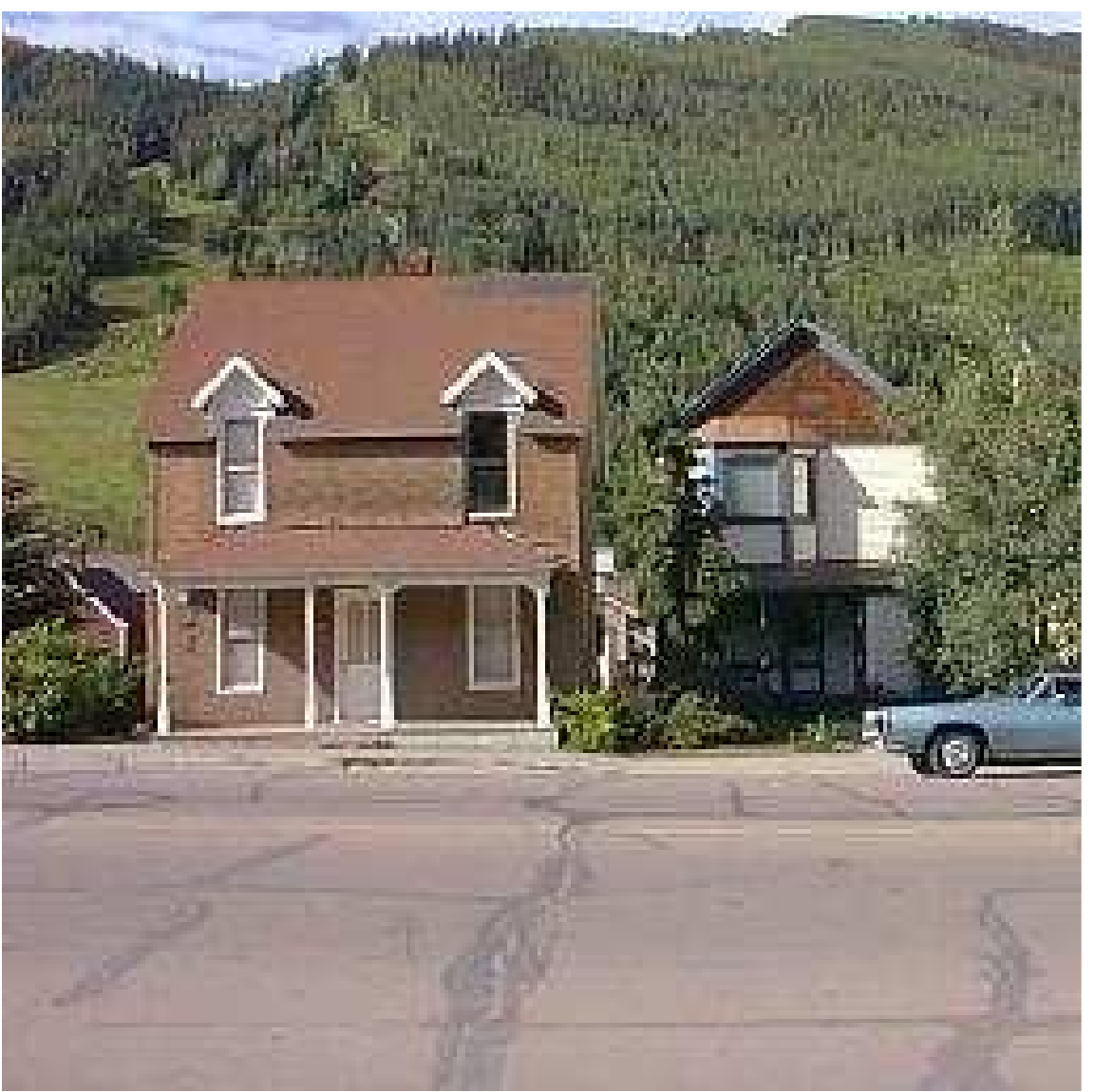}}
 \subfloat{\epsfysize 2.5cm \epsfxsize 2.5cm
 \leavevmode
 \epsffile{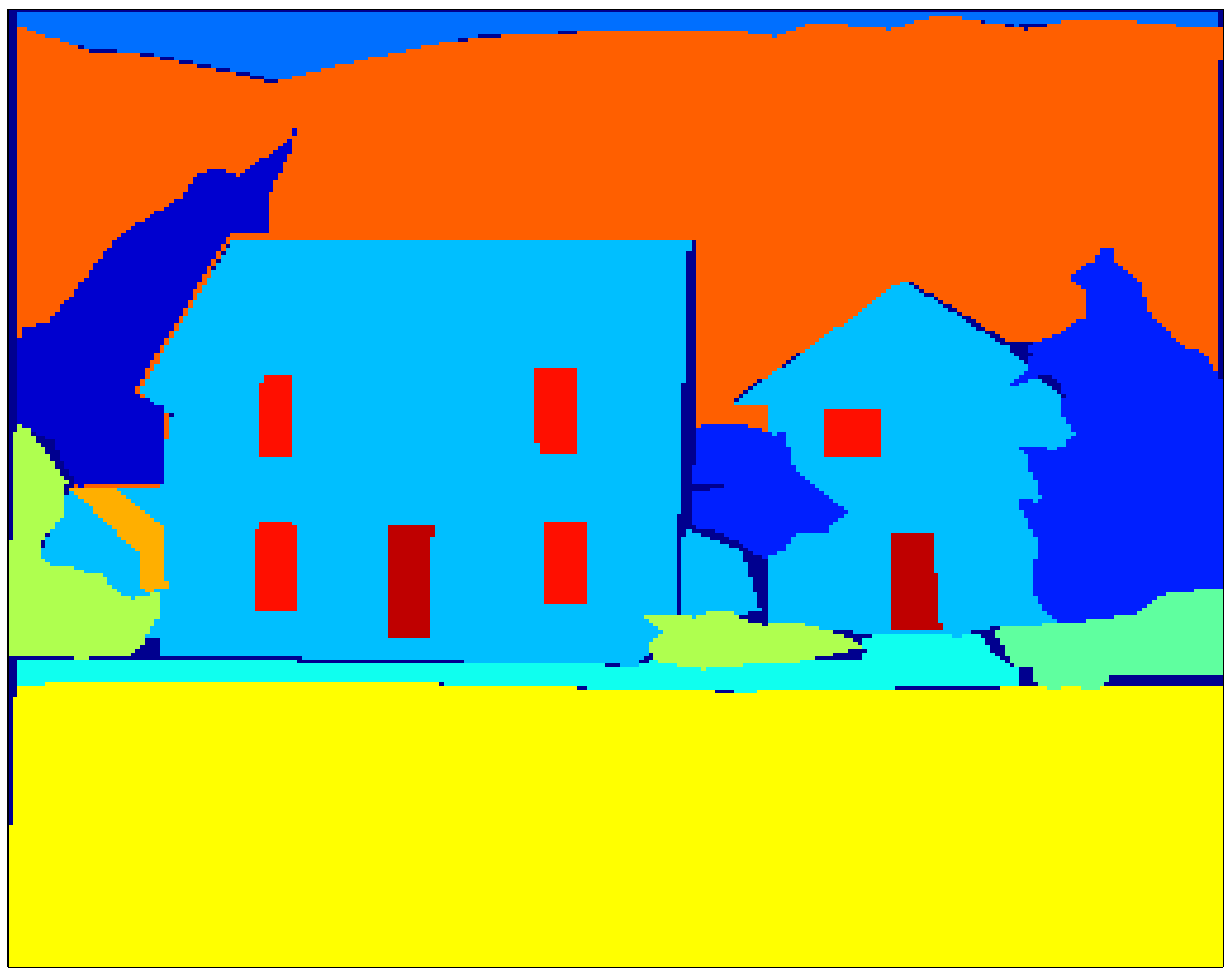}}
 \subfloat{\epsfysize 2.5cm \epsfxsize 2.5cm
 \leavevmode
 \epsffile{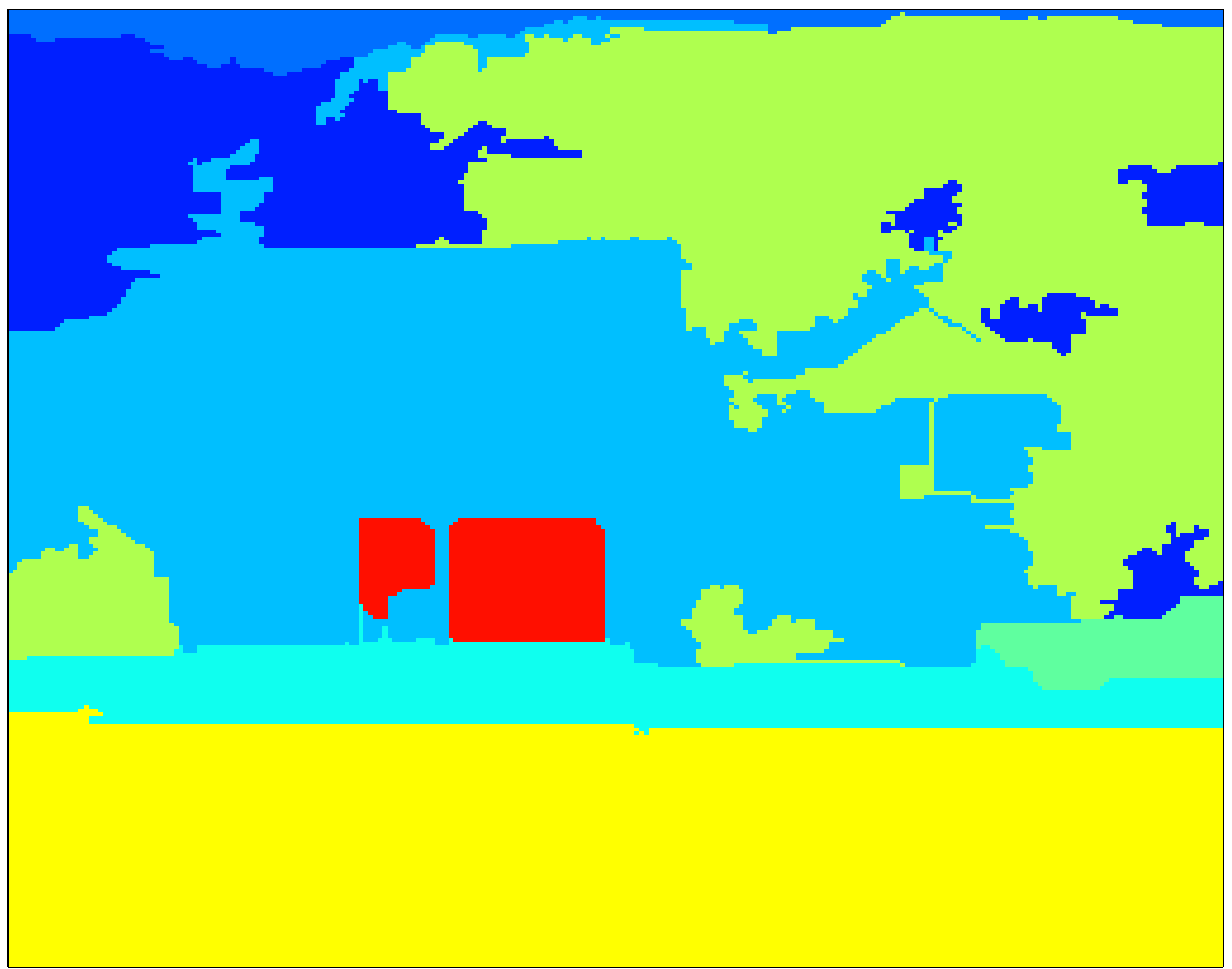}}
\subfloat{\epsfysize 2.5cm \epsfxsize 2.5cm
 \leavevmode
 \epsffile{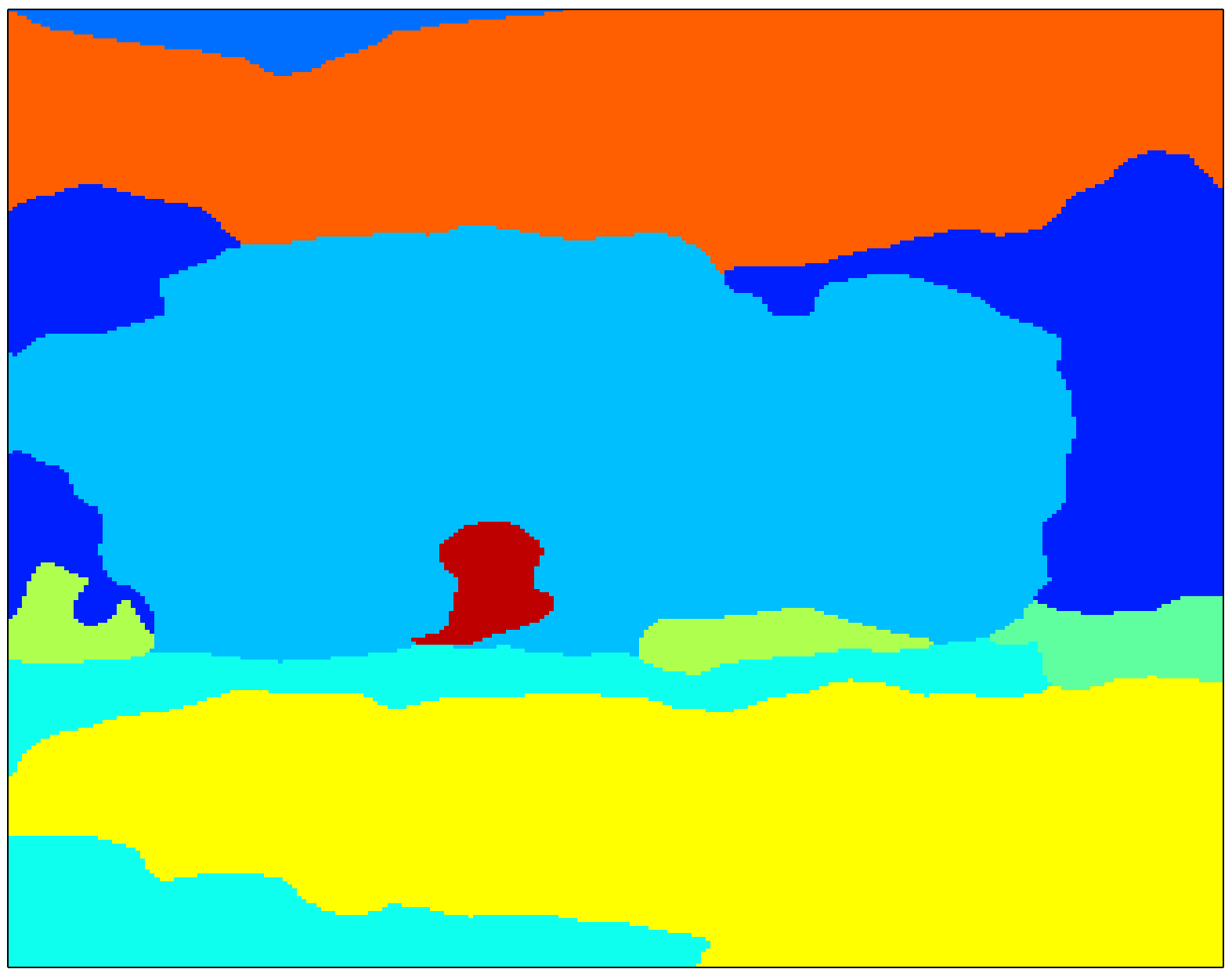}}
  \subfloat{\epsfysize 2.5cm \epsfxsize 2.5cm
 \leavevmode
 \epsffile{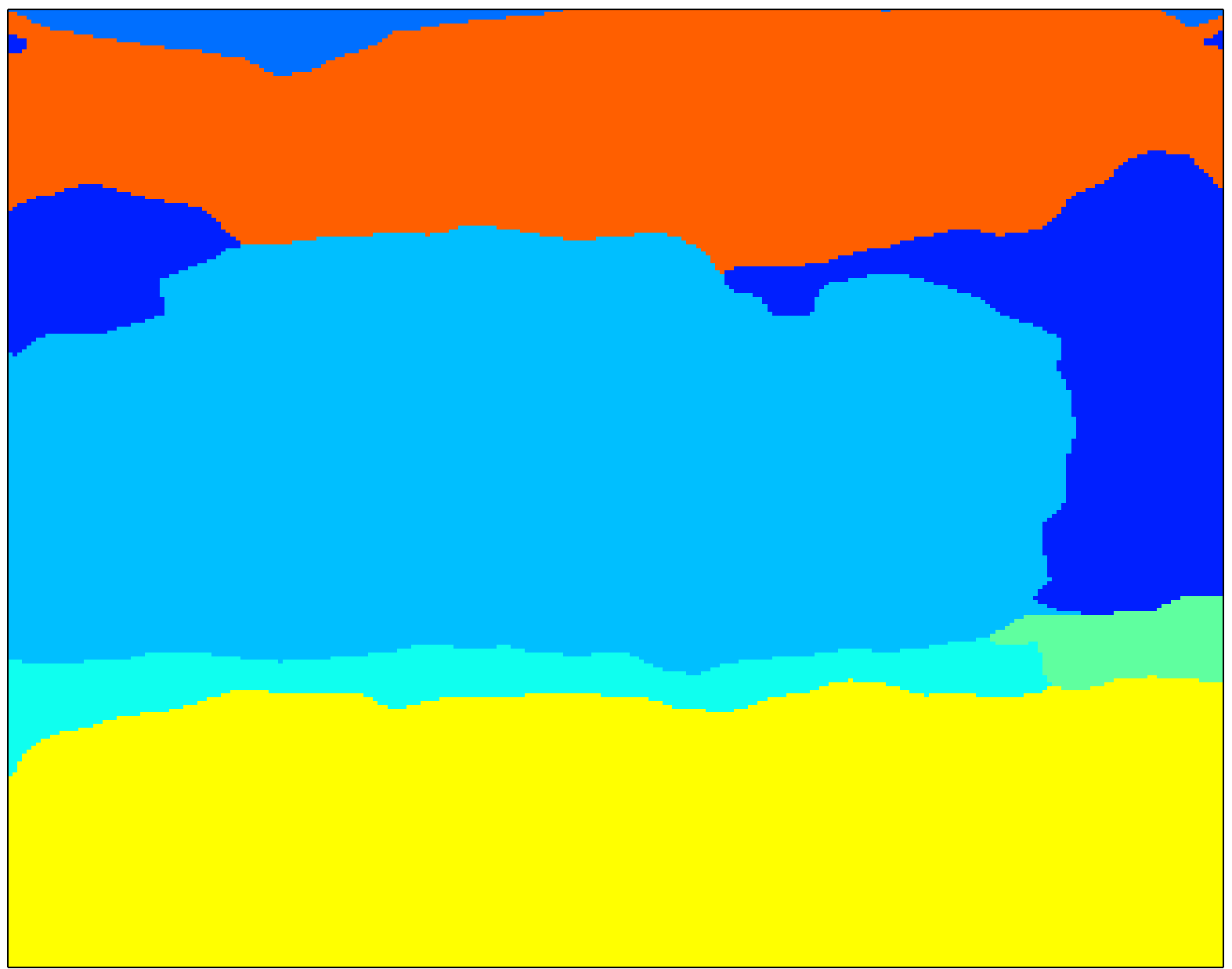}}
  \subfloat{\epsfysize 2.5cm \epsfxsize 1.25cm
 \leavevmode
 \epsffile{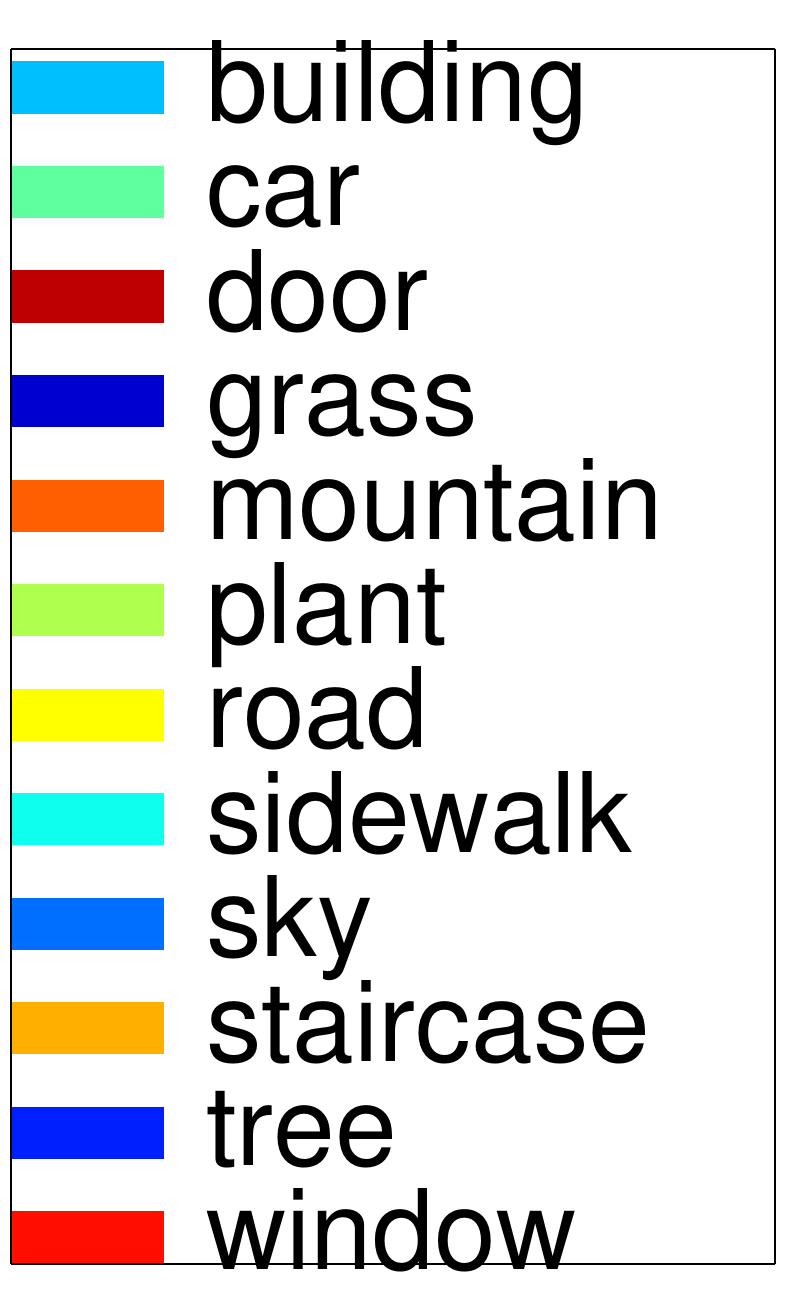}}

\setcounter{subfigure}{0}
\centering \subfloat[Part1][]{\epsfysize 2.5cm \epsfxsize 2.5cm
 \leavevmode
 \epsffile{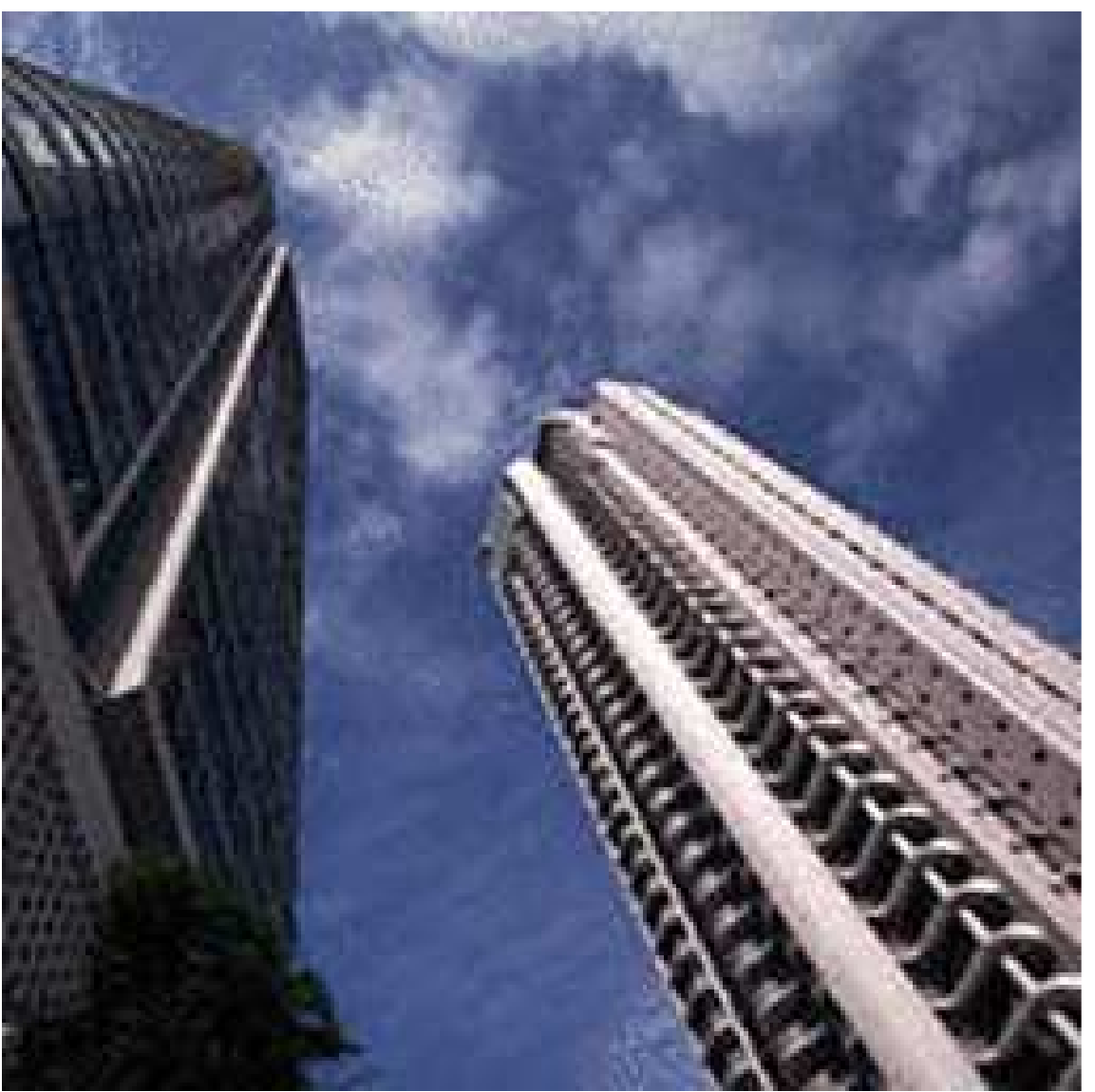}}
 \subfloat[Part2][]{\epsfysize 2.5cm \epsfxsize 2.5cm
 \leavevmode
 \epsffile{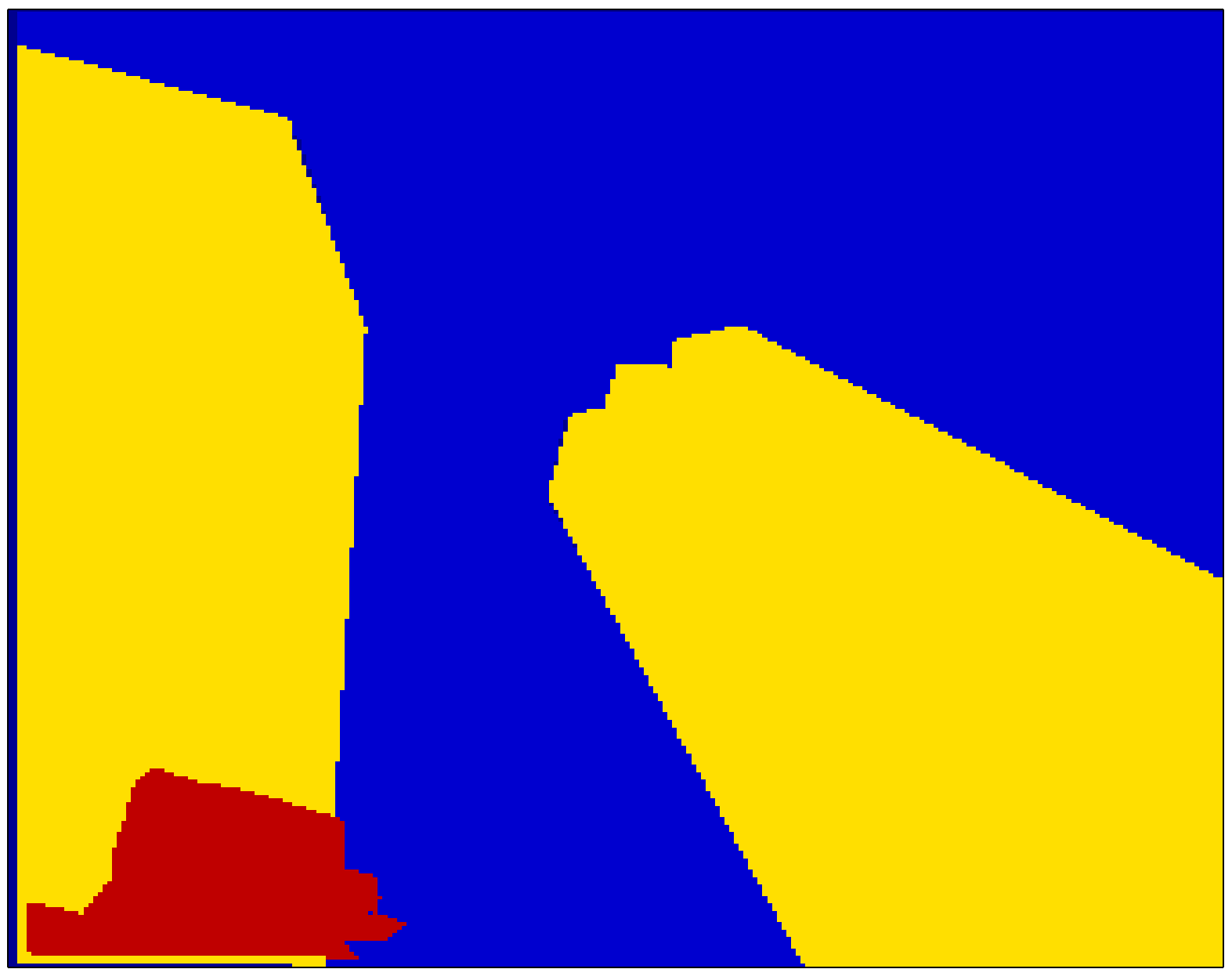}}
 \subfloat[Part3][]{\epsfysize 2.5cm \epsfxsize 2.5cm
 \leavevmode
 \epsffile{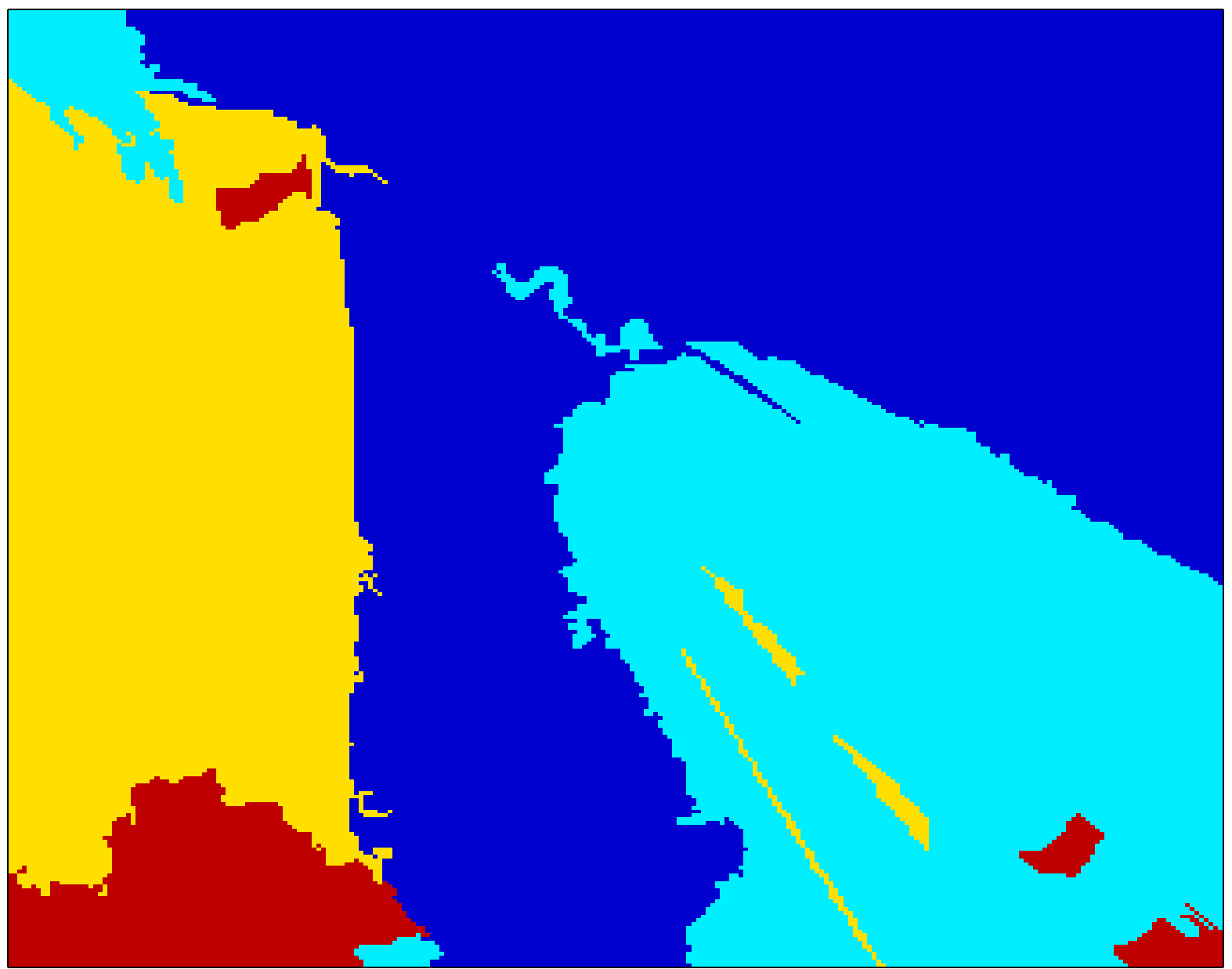}}
\subfloat[Part4][]{\epsfysize 2.5cm \epsfxsize 2.5cm
 \leavevmode
 \epsffile{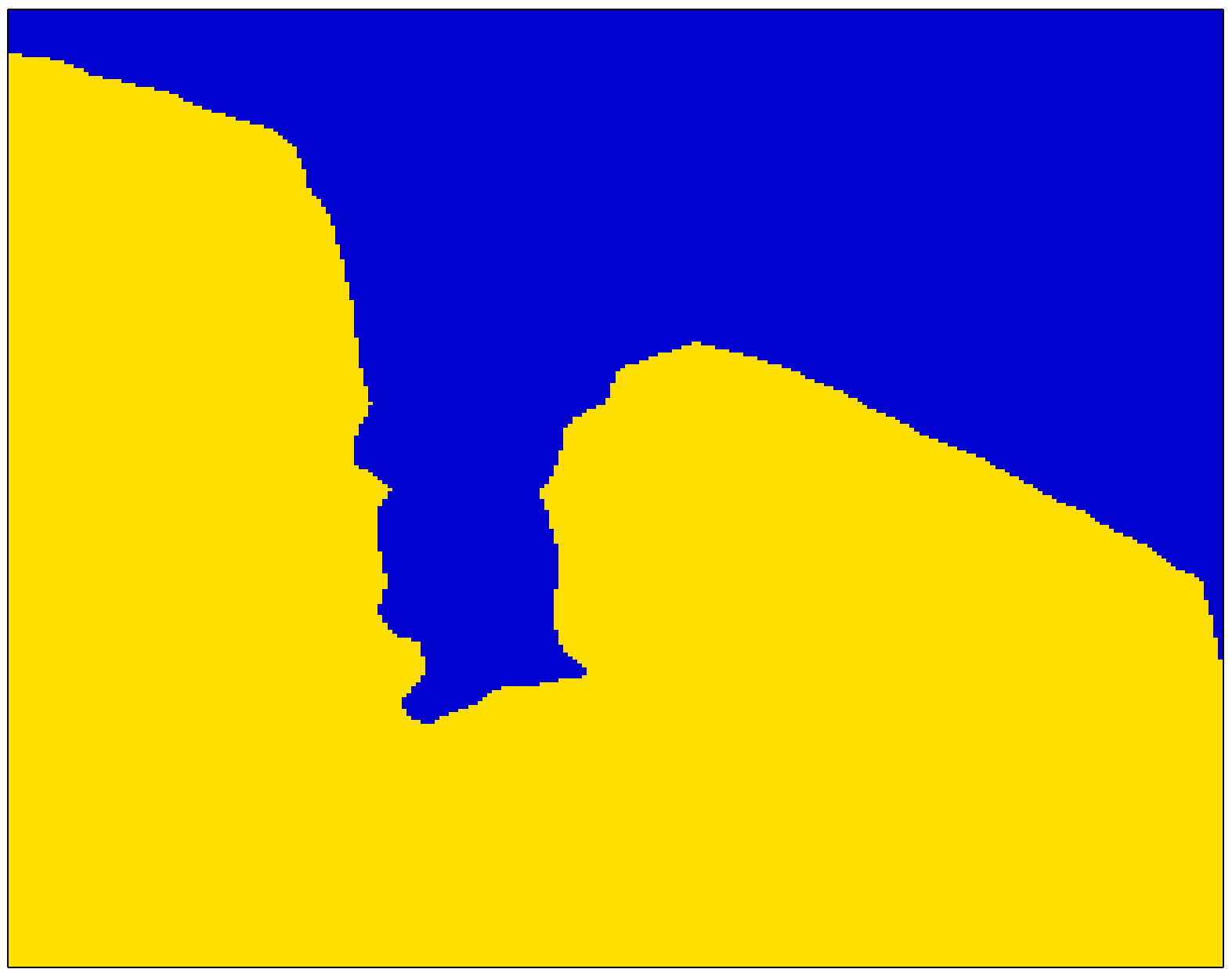}}
  \subfloat[Part5][]{\epsfysize 2.5cm \epsfxsize 2.5cm
 \leavevmode
  \includegraphics[width=2.5cm, height=2.5cm]{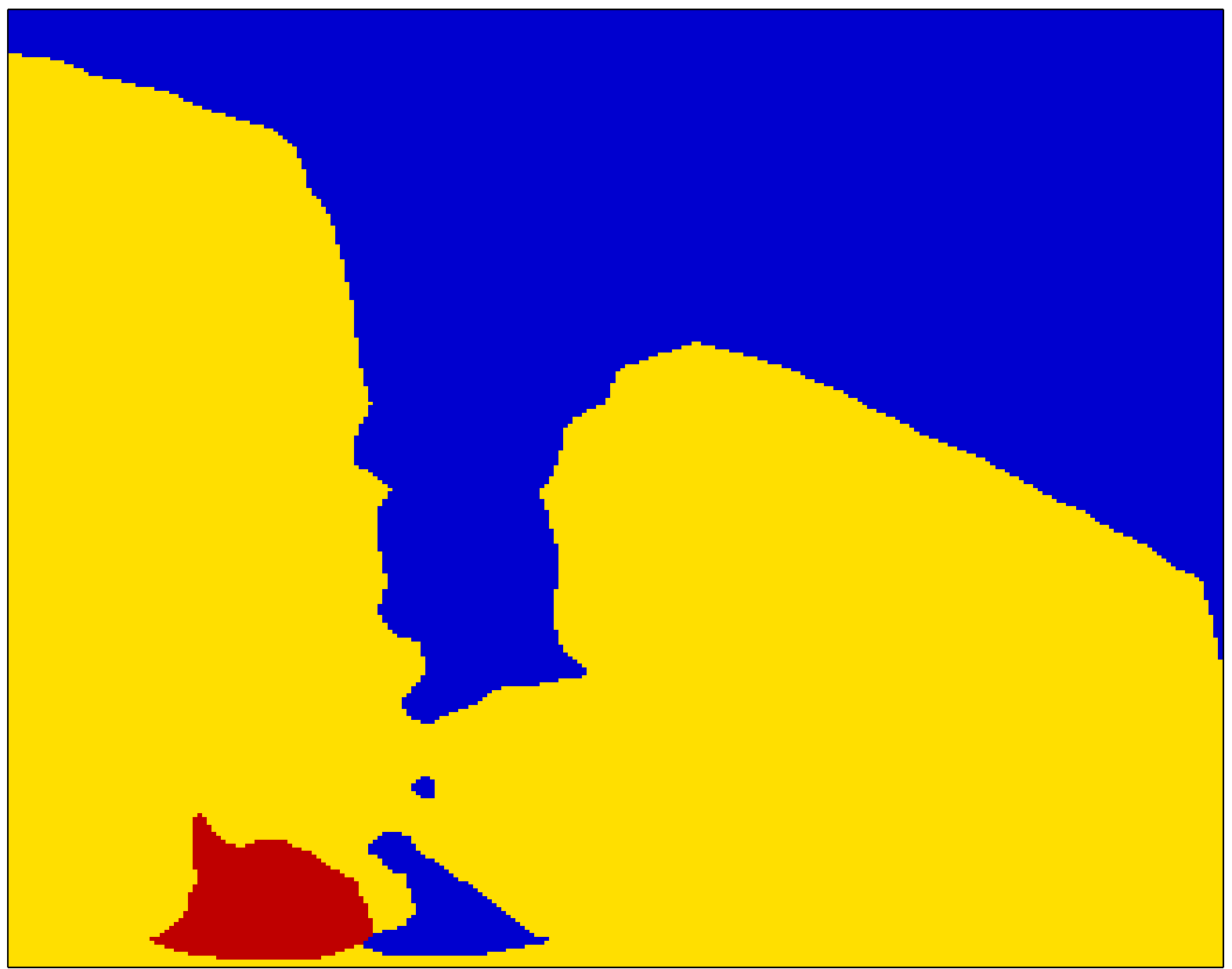}}
  \subfloat{\epsfysize 2.5cm \epsfxsize 1.25cm
 \leavevmode
 \epsffile{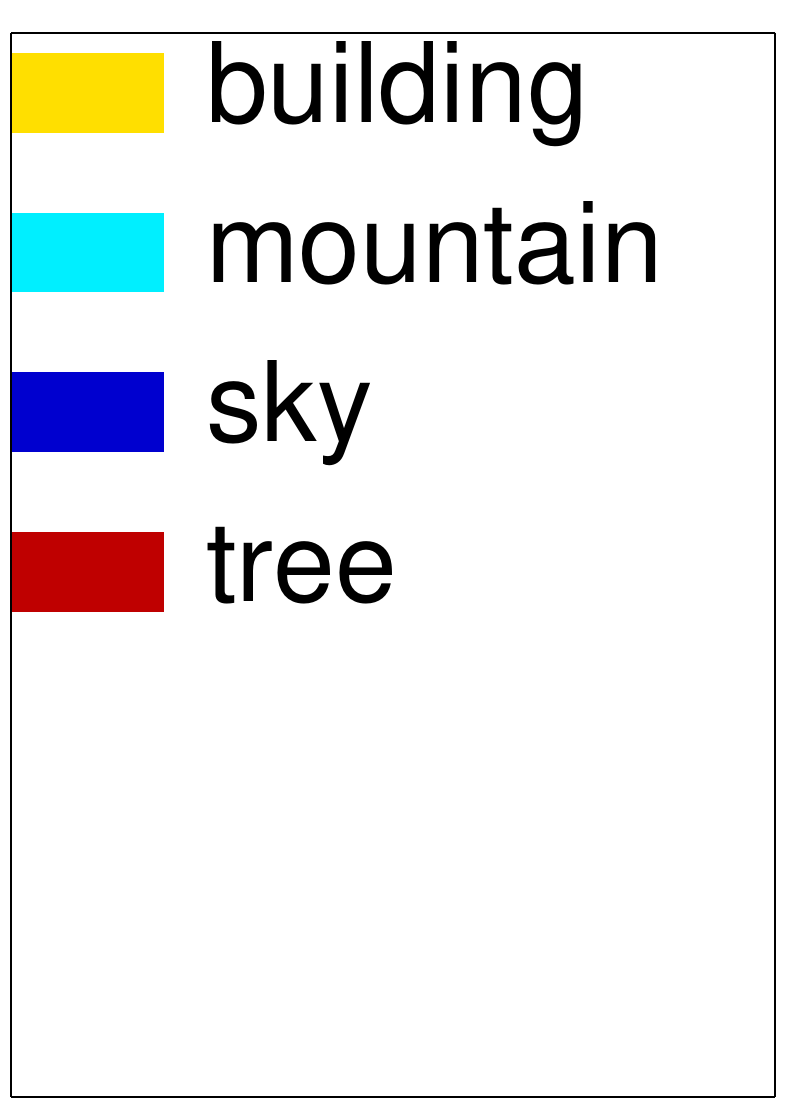}}
\caption{Visual comparison: (a) Original image; (b) Ground truth;  (c) Seg1+MRF ; (d) Seg2+MRF ; (e) DC2 (Seg1,Seg2)}
\label{parse_result}
\end{figure*}

19-class LabelMe dataset contains 350 images with 19 classes (such as {\it tree}, {\it field}, {\it building}, {\it rock}, etc.). The dataset is split into 250 training images and 100 test images. The retrieval set size is set at 50 images, due to  the smaller size of training set. KCB encoding is used for SIFT features. FCN-8s network architecture is transferred from SIFT Flow, adapted for 19-class evaluation and re-trained with the given 250 image LabelMe training set. PSP architecture is not tested due to the small size of training set.


Table \ref{table:accuracyLMS} reports the pixel-level and average per-class labeling accuracies of (Seg1,Seg2) using DC1 and DC2 for LabelMe. Results from literature are provided for \cite{nguyen_cnn}, \cite{adapt_super}, \cite{Myeong2013} and our initial work in \cite{Ates2017}. Our proposed algorithms surpass the state-of-art in this dataset, improving pixel accuracy by 4.9\% and mean class accuracy by 12.9\% over \cite{nguyen_cnn}. We also outperform FCN-8s result by 3.5\% and 6.1\% in pixel and mean class accuracies, respectively.

For LabelMe dataset, DC1 is better than DC2 by 0.5\% in pixel accuracy, but DC2 gives higher mean class accuracy. DC1 improves the pixel and mean class accuracies of Seg2+MRF by 1.4\% and 3.0\%, respectively. This result indicates the importance of inter- neighborhoods in combining the complementary information of alternative methods. As opposed to SIFT Flow, (Seg1,Seg2) provides improvement in mean class accuracy in this dataset. This implies that, when both Seg1 and Seg2 have comparable performance, our multi-hypothesis MRF framework boosts not only the pixel accuracies but also the mean class accuracies of the tested methods.


At Figure \ref{parse_result}, parsing results of Seg1+MRF, Seg2+MRF and DC2(Seg1,Seg2) are compared visually for some selected test images from SIFT Flow. DC2 labelings are generally more consistent and accurate than those of both Seg1+MRF and Seg2+MRF. In the top figure, {\it field} is correctly identified, even though both tested methods label the area as {\it grass}. In the other three figures, and typically throughout the SIFT Flow dataset, DC2 outcome is at least as good as the better result of Seg1+MRF and Seg2+MRF. In other words, the proposed approach manages to select correctly the more probable label between the two hypotheses by making use of the intricate contextual constraints in intra- and inter- neighborhoods.

\subsection{Computational Cost}
The time complexity of the proposed approach is dominated by the running times of individual parsing algorithms. MRF optimization is carried out by the fast $\alpha$-expansion method of \cite{Min_cut}. The computational cost of this optimization is proportional to the total number of superpixels in $SP_1$, $SP_2$ and $SP_3$.

Simulations are carried out on a single PC with 4-core CPU at 3.70 GHz and 16 GB RAM. {\it Caffe} implementations of FCN and PSP are used for testing.
For SIFT Flow dataset, FCN and PSP take 2.2 and 5.9 secs, respectively, per image. SuperParsing labels images in 8 secs on average, using its unoptimized and un-parallelized MATLAB implementation. Multi-hypothesis MRF model optimization takes merely 0.25 secs on average, using unoptimized MATLAB interface. Hence the proposed framework provides fast inference for contextual modeling and fusion of parsing algorithms with different levels of performance and complexity.

\section{Conclusion}
\label{sonuc}
In this paper a novel contextual modeling framework  is introduced for semantic scene segmentation. This framework defines contextual constraints over inter- and intra- neighborhoods for multiple segmentations of the same image. In addition to producing spatially more consistent parsing results, the proposed approach carries out labeling at a finer scale over the intersecting regions of alternative segmentations. We have shown that, when both alternatives have comparable labeling performance, our contextual models improve both the pixel and the mean class accuracies of tested methods. We have used this framework as a post-processing step at the outputs of deep FCN-8s and PSP architectures and obtained state-of-the-art parsing results in two well-known datasets.

As future work, we plan to advance our contextual inference approach using more advanced data cost and smoothness models. Other CNN architectures and superpixel-based parsing methods could be tested within the given framework. Also this MRF framework could be integrated into the CNN architecture and the model parameters could be learned using end-to-end training of the whole system.


\section*{Acknowledgments}
This work is supported in part by TUBITAK project no: 115E307 and by Isik University BAP project no: 14A205.

\bibliographystyle{model2-names}
\bibliography{refs_prletter_2017}

\begin{thebibliography}{38}
\expandafter\ifx\csname natexlab\endcsname\relax\def\natexlab#1{#1}\fi
\providecommand{\url}[1]{\texttt{#1}}
\providecommand{\href}[2]{#2}
\providecommand{\path}[1]{#1}
\providecommand{\DOIprefix}{doi:}
\providecommand{\ArXivprefix}{arXiv:}
\providecommand{\URLprefix}{URL: }
\providecommand{\Pubmedprefix}{pmid:}
\providecommand{\doi}[1]{\href{http://dx.doi.org/#1}{\path{#1}}}
\providecommand{\Pubmed}[1]{\href{pmid:#1}{\path{#1}}}
\providecommand{\bibinfo}[2]{#2}
\ifx\xfnm\relax \def\xfnm[#1]{\unskip,\space#1}\fi
\bibitem[{Ak and Ates(2015)}]{multi_hypothesis}
\bibinfo{author}{Ak, K.E.}, \bibinfo{author}{Ates, H.F.}, \bibinfo{year}{2015}.
\newblock \bibinfo{title}{Scene segmentation and labeling using
  multi-hypothesis superpixels}, in: \bibinfo{booktitle}{Signal Process. and
  Comm. Appl. Conf. (SIU)}, pp. \bibinfo{pages}{847--850}.
\bibitem[{Ates and Sunetci(2017)}]{Ates2017}
\bibinfo{author}{Ates, H.F.}, \bibinfo{author}{Sunetci, S.},
  \bibinfo{year}{2017}.
\newblock \bibinfo{title}{Improving semantic segmentation with generalized
  models of local context}, in: \bibinfo{booktitle}{17th Int. Conf. Computer
  Analysis Images Patterns (CAIP)}, pp. \bibinfo{pages}{320--330}.
\bibitem[{Bell et~al.(2015)Bell, Upchurch, Snavely and Bala}]{crf1}
\bibinfo{author}{Bell, S.}, \bibinfo{author}{Upchurch, P.},
  \bibinfo{author}{Snavely, N.}, \bibinfo{author}{Bala, K.},
  \bibinfo{year}{2015}.
\newblock \bibinfo{title}{Material recognition in the wild with the materials
  in context database}, in: \bibinfo{booktitle}{Proc. IEEE Conf. Comp. Vision
  Pattern Recog. (CVPR)}, pp. \bibinfo{pages}{3479--3487}.
\bibitem[{Boykov and Kolmogorov(2004)}]{Min_cut}
\bibinfo{author}{Boykov, Y.}, \bibinfo{author}{Kolmogorov, V.},
  \bibinfo{year}{2004}.
\newblock \bibinfo{title}{An experimental comparison of min-cut/max-flow
  algorithms for energy minimization in vision}.
\newblock \bibinfo{journal}{IEEE Trans. Pattern Anal. Mach. Intell}
  \bibinfo{volume}{26}, \bibinfo{pages}{1124--1137}.
\bibitem[{Byeon et~al.(2015)Byeon, Breuel, Raue and Liwicki}]{rnn1}
\bibinfo{author}{Byeon, W.}, \bibinfo{author}{Breuel, T.M.},
  \bibinfo{author}{Raue, F.}, \bibinfo{author}{Liwicki, M.},
  \bibinfo{year}{2015}.
\newblock \bibinfo{title}{Scene labeling with lstm recurrent neural networks},
  in: \bibinfo{booktitle}{Proc. IEEE Conf. Computer Vision Pattern Recog.}, pp.
  \bibinfo{pages}{3547--3555}.
\bibitem[{Chatfield et~al.(2014)Chatfield, Simonyan, Vedaldi and
  Zisserman}]{Chatfield14}
\bibinfo{author}{Chatfield, K.}, \bibinfo{author}{Simonyan, K.},
  \bibinfo{author}{Vedaldi, A.}, \bibinfo{author}{Zisserman, A.},
  \bibinfo{year}{2014}.
\newblock \bibinfo{title}{Return of the devil in the details: Delving deep into
  convolutional nets}, in: \bibinfo{booktitle}{Proc. British Machine Vision
  Conf.}
\bibitem[{Cheng et~al.(2017)Cheng, He and Zhang}]{stacked}
\bibinfo{author}{Cheng, F.}, \bibinfo{author}{He, X.}, \bibinfo{author}{Zhang,
  H.}, \bibinfo{year}{2017}.
\newblock \bibinfo{title}{Stacked learning to search for scene labeling}.
\newblock \bibinfo{journal}{IEEE Trans. Image Process.} \bibinfo{volume}{26},
  \bibinfo{pages}{1887--1898}.
\bibitem[{Deng et~al.(2009)Deng, Dong, Socher, Li, Li and Fei-Fei}]{imagenet}
\bibinfo{author}{Deng, J.}, \bibinfo{author}{Dong, W.},
  \bibinfo{author}{Socher, R.}, \bibinfo{author}{Li, L.J.},
  \bibinfo{author}{Li, K.}, \bibinfo{author}{Fei-Fei, L.},
  \bibinfo{year}{2009}.
\newblock \bibinfo{title}{Imagenet: A large-scale hierarchical image database},
  in: \bibinfo{booktitle}{Proc. IEEE Conf. Comp. Vision Pattern Recog. (CVPR)},
  pp. \bibinfo{pages}{248--255}.
\bibitem[{Dong et~al.(2016)Dong, Feng and Zhang}]{dong}
\bibinfo{author}{Dong, L.}, \bibinfo{author}{Feng, N.}, \bibinfo{author}{Zhang,
  Q.}, \bibinfo{year}{2016}.
\newblock \bibinfo{title}{Lsi: Latent semantic inference for natural image
  segmentation}.
\newblock \bibinfo{journal}{Pattern Recognition} \bibinfo{volume}{59},
  \bibinfo{pages}{282--291}.
\bibitem[{Eigen and Fergus(2015)}]{ms_eigen}
\bibinfo{author}{Eigen, D.}, \bibinfo{author}{Fergus, R.},
  \bibinfo{year}{2015}.
\newblock \bibinfo{title}{Predicting depth, surface normals and semantic labels
  with a common multi-scale convolutional architecture}, in:
  \bibinfo{booktitle}{Proc. IEEE Conf. Computer Vision}, pp.
  \bibinfo{pages}{2650--2658}.
\bibitem[{Farabet et~al.(2013)Farabet, Couprie, Najman and
  LeCun}]{multiscale_parse}
\bibinfo{author}{Farabet, C.}, \bibinfo{author}{Couprie, C.},
  \bibinfo{author}{Najman, L.}, \bibinfo{author}{LeCun, Y.},
  \bibinfo{year}{2013}.
\newblock \bibinfo{title}{Learning hierarchical features for scene labeling}.
\newblock \bibinfo{journal}{IEEE Trans. Pattern Anal. Mach. Intell.}
  \bibinfo{volume}{35}, \bibinfo{pages}{1915--1929}.
\bibitem[{Felzenszwalb and Huttenlocher(2004)}]{graph}
\bibinfo{author}{Felzenszwalb, P.F.}, \bibinfo{author}{Huttenlocher, D.P.},
  \bibinfo{year}{2004}.
\newblock \bibinfo{title}{Efficient graph-based image segmentation}.
\newblock \bibinfo{journal}{Int. Journal Computer Vision} \bibinfo{volume}{59},
  \bibinfo{pages}{167--181}.
\bibitem[{van Gemert et~al.(2008)van Gemert, Geusebroek, Veenman and
  Smeulders}]{KCB}
\bibinfo{author}{van Gemert, J.C.}, \bibinfo{author}{Geusebroek, J.M.},
  \bibinfo{author}{Veenman, C.J.}, \bibinfo{author}{Smeulders, A.W.M.},
  \bibinfo{year}{2008}.
\newblock \bibinfo{title}{Kernel codebooks for scene categorization}, in:
  \bibinfo{booktitle}{Proc. European Conf. Computer Vision}, pp.
  \bibinfo{pages}{696--709}.
\bibitem[{George(2015)}]{fischer}
\bibinfo{author}{George, M.}, \bibinfo{year}{2015}.
\newblock \bibinfo{title}{Image parsing with a wide range of classes and
  scene-level context}, in: \bibinfo{booktitle}{Proc. IEEE Conf. Comp. Vision
  Pattern Recog. (CVPR)}, pp. \bibinfo{pages}{3622--3630}.
\bibitem[{Jain et~al.(2010)Jain, Gupta and Davis}]{labelmesub}
\bibinfo{author}{Jain, A.}, \bibinfo{author}{Gupta, A.},
  \bibinfo{author}{Davis, L.}, \bibinfo{year}{2010}.
\newblock \bibinfo{title}{Learning what and how of contextual models for scene
  labeling}, in: \bibinfo{booktitle}{Proc. European Conf. Computer Vision}, pp.
  \bibinfo{pages}{199--212}.
\bibitem[{Krizhevsky et~al.(2012)Krizhevsky, Sutskever and
  Hinton}]{NIPS2012_4824}
\bibinfo{author}{Krizhevsky, A.}, \bibinfo{author}{Sutskever, I.},
  \bibinfo{author}{Hinton, G.E.}, \bibinfo{year}{2012}.
\newblock \bibinfo{title}{Imagenet classification with deep convolutional
  neural networks}, in: \bibinfo{booktitle}{Advances in Neural Information
  Processing Systems 25}, pp. \bibinfo{pages}{1097--1105}.
\bibitem[{Ladicky et~al.(2014)Ladicky, Russell, Kohli and Torr}]{ahrf}
\bibinfo{author}{Ladicky, L.}, \bibinfo{author}{Russell, C.},
  \bibinfo{author}{Kohli, P.}, \bibinfo{author}{Torr, P.H.S.},
  \bibinfo{year}{2014}.
\newblock \bibinfo{title}{Associative hierarchical random fields}.
\newblock \bibinfo{journal}{IEEE Trans. Pattern Anal. Mach. Intell.}
  \bibinfo{volume}{36}, \bibinfo{pages}{1056--1077}.
\bibitem[{Li et~al.(2016)Li, Gan, Liang, Yu, Cheng and Lin}]{rnn2}
\bibinfo{author}{Li, Z.}, \bibinfo{author}{Gan, Y.}, \bibinfo{author}{Liang,
  X.}, \bibinfo{author}{Yu, Y.}, \bibinfo{author}{Cheng, H.},
  \bibinfo{author}{Lin, L.}, \bibinfo{year}{2016}.
\newblock \bibinfo{title}{Lstm-cf: Unifying context modeling and fusion with
  lstms for rgb-d scene labeling}, in: \bibinfo{booktitle}{Proc. European Conf.
  Computer Vision}, pp. \bibinfo{pages}{541--557}.
\bibitem[{Liang et~al.(2015)Liang, Hu and Zhang}]{rcnn}
\bibinfo{author}{Liang, M.}, \bibinfo{author}{Hu, X.}, \bibinfo{author}{Zhang,
  B.}, \bibinfo{year}{2015}.
\newblock \bibinfo{title}{Convolutional neural networks with intra-layer
  recurrent connections for scene labeling}, in: \bibinfo{booktitle}{Advances
  in Neural Information Processing Systems}, pp. \bibinfo{pages}{937--945}.
\bibitem[{Liang{-}Chieh et~al.(2018)Liang{-}Chieh, Papandreou, Kokkinos, Murphy
  and Yuille}]{deeplab}
\bibinfo{author}{Liang{-}Chieh, C.}, \bibinfo{author}{Papandreou, G.},
  \bibinfo{author}{Kokkinos, I.}, \bibinfo{author}{Murphy, K.},
  \bibinfo{author}{Yuille, A.L.}, \bibinfo{year}{2018}.
\newblock \bibinfo{title}{Deeplab: Semantic image segmentation with deep
  convolutional nets, atrous convolution, and fully connected crfs}.
\newblock \bibinfo{journal}{IEEE Trans. Pattern Anal. Mach. Intell.}
  \bibinfo{volume}{40}, \bibinfo{pages}{834--848}.
\bibitem[{Liu et~al.(2011)Liu, Yuen and Torralba}]{siftflow}
\bibinfo{author}{Liu, C.}, \bibinfo{author}{Yuen, J.},
  \bibinfo{author}{Torralba, A.}, \bibinfo{year}{2011}.
\newblock \bibinfo{title}{{SIFT} {Flow}: Dense correspondence across scenes and
  its applications}.
\newblock \bibinfo{journal}{IEEE Trans. Pattern Anal. Mach. Intell.}
  \bibinfo{volume}{33}, \bibinfo{pages}{978--994}.
\bibitem[{Liu et~al.(2016)Liu, Rabinovich and Berg.}]{parsenet}
\bibinfo{author}{Liu, W.}, \bibinfo{author}{Rabinovich, A.},
  \bibinfo{author}{Berg., A.C.}, \bibinfo{year}{2016}.
\newblock \bibinfo{title}{Parsenet: Looking wider to see better}, in:
  \bibinfo{booktitle}{Proc. Int. Conf. Learning Represent. (ICLR) Workshop}.
\bibitem[{Morales-Gonzalez et~al.(2018)Morales-Gonzalez, Garcia-Reyes and
  Sucar}]{morales}
\bibinfo{author}{Morales-Gonzalez, A.}, \bibinfo{author}{Garcia-Reyes, E.},
  \bibinfo{author}{Sucar, L.E.}, \bibinfo{year}{2018}.
\newblock \bibinfo{title}{Image annotation by a hierarchical and iterative
  combination of recognition and segmentation}.
\newblock \bibinfo{journal}{Multimedia Tools and Applications}
  \bibinfo{volume}{32}, \bibinfo{pages}{1860014}.
\bibitem[{Myeong and Lee(2013)}]{Myeong2013}
\bibinfo{author}{Myeong, H.}, \bibinfo{author}{Lee, K.}, \bibinfo{year}{2013}.
\newblock \bibinfo{title}{Tensor-based high-order semantic relation transfer
  for semantic scene segmentation}, in: \bibinfo{booktitle}{Proc. IEEE Conf.
  Comp. Vision Pattern Recog. (CVPR)}, pp. \bibinfo{pages}{3073--3080}.
\bibitem[{Nguyen et~al.(2016)Nguyen, Liu and Nguyen}]{nguyen_cnn}
\bibinfo{author}{Nguyen, T.V.}, \bibinfo{author}{Liu, L.},
  \bibinfo{author}{Nguyen, K.}, \bibinfo{year}{2016}.
\newblock \bibinfo{title}{Exploiting generic multi-level convolutional neural
  networks for scene understanding}, in: \bibinfo{booktitle}{Proc. 14th Int.
  Conf. Control Automation Robotics Vision (ICARCV)}, pp.
  \bibinfo{pages}{1--6}.
\bibitem[{Nguyen et~al.(2015)Nguyen, Lu, Sepulveda and Yan}]{adapt_super}
\bibinfo{author}{Nguyen, T.V.}, \bibinfo{author}{Lu, C.},
  \bibinfo{author}{Sepulveda, J.}, \bibinfo{author}{Yan, S.},
  \bibinfo{year}{2015}.
\newblock \bibinfo{title}{Adaptive nonparametric image parsing}.
\newblock \bibinfo{journal}{IEEE Trans. Circuits Syst. Video Tech.}
  \bibinfo{volume}{25}, \bibinfo{pages}{1565--1575}.
\bibitem[{Pantofaru et~al.(2008)Pantofaru, Schmid and Hebert}]{pantofaru}
\bibinfo{author}{Pantofaru, C.}, \bibinfo{author}{Schmid, C.},
  \bibinfo{author}{Hebert, M.}, \bibinfo{year}{2008}.
\newblock \bibinfo{title}{Object recognition by integrating multiple image
  segmentations}, in: \bibinfo{booktitle}{Proc. European Conf. Computer
  Vision}, pp. \bibinfo{pages}{481--494}.
\bibitem[{Shelhamer et~al.(2016)Shelhamer, Long and Darrell}]{fcnn}
\bibinfo{author}{Shelhamer, E.}, \bibinfo{author}{Long, J.},
  \bibinfo{author}{Darrell, T.}, \bibinfo{year}{2016}.
\newblock \bibinfo{title}{Fully convolutional networks for semantic
  segmentation}.
\newblock \bibinfo{journal}{IEEE Trans. Pattern Anal. Mach. Intell.}
  \bibinfo{volume}{39}, \bibinfo{pages}{640--651}.
\bibitem[{Shuai et~al.(2016)Shuai, Zuo, Wang and Wang}]{dag_rcnn}
\bibinfo{author}{Shuai, B.}, \bibinfo{author}{Zuo, Z.}, \bibinfo{author}{Wang,
  B.}, \bibinfo{author}{Wang, G.}, \bibinfo{year}{2016}.
\newblock \bibinfo{title}{Dag-recurrent neural networks for scene labeling},
  in: \bibinfo{booktitle}{Proc. IEEE Conf. Comp. Vision Pattern Recog. (CVPR)},
  pp. \bibinfo{pages}{3620--3629}.
\bibitem[{Tighe and Lazebnik(2013)}]{superparsing}
\bibinfo{author}{Tighe, J.}, \bibinfo{author}{Lazebnik, S.},
  \bibinfo{year}{2013}.
\newblock \bibinfo{title}{Superparsing: scalable nonparametric image parsing
  with superpixels}.
\newblock \bibinfo{journal}{Int. Journal Computer Vision}
  \bibinfo{volume}{101}, \bibinfo{pages}{329--349}.
\bibitem[{Tighe et~al.(2015)Tighe, Niethammer and Lazebnik}]{Tighe2015}
\bibinfo{author}{Tighe, J.}, \bibinfo{author}{Niethammer, M.},
  \bibinfo{author}{Lazebnik, S.}, \bibinfo{year}{2015}.
\newblock \bibinfo{title}{Scene parsing with object instance inference using
  regions and per-exemplar detectors}.
\newblock \bibinfo{journal}{Int. Journal Computer Vision}
  \bibinfo{volume}{112}, \bibinfo{pages}{150--171}.
\bibitem[{Vieux et~al.(2012)Vieux, Benois-Pineau, Domenger and
  Braquelaire}]{vieux}
\bibinfo{author}{Vieux, R.}, \bibinfo{author}{Benois-Pineau, J.},
  \bibinfo{author}{Domenger, J.P.}, \bibinfo{author}{Braquelaire, A.},
  \bibinfo{year}{2012}.
\newblock \bibinfo{title}{Segmentation-based multi-class semantic object
  detection}.
\newblock \bibinfo{journal}{Multimedia Tools and Applications}
  \bibinfo{volume}{60}, \bibinfo{pages}{305--326}.
\bibitem[{Wang et~al.(2010)Wang, Yang, Yu, Lv, Huang and Gong}]{LLC}
\bibinfo{author}{Wang, J.}, \bibinfo{author}{Yang, J.}, \bibinfo{author}{Yu,
  K.}, \bibinfo{author}{Lv, F.}, \bibinfo{author}{Huang, T.},
  \bibinfo{author}{Gong, Y.}, \bibinfo{year}{2010}.
\newblock \bibinfo{title}{Locality-constrained linear coding for image
  classification}, in: \bibinfo{booktitle}{Proc. IEEE Conf. Comp. Vision
  Pattern Recog. (CVPR)}, pp. \bibinfo{pages}{3360--3367}.
\bibitem[{Yang et~al.(2014)Yang, Price, Cohen and Yang}]{context_rare}
\bibinfo{author}{Yang, J.}, \bibinfo{author}{Price, B.},
  \bibinfo{author}{Cohen, S.}, \bibinfo{author}{Yang, M.H.},
  \bibinfo{year}{2014}.
\newblock \bibinfo{title}{Context driven scene parsing with attention to rare
  classes}, in: \bibinfo{booktitle}{Proc. IEEE Conf. Comp. Vision Pattern
  Recog. (CVPR)}, pp. \bibinfo{pages}{3294--3301}.
\bibitem[{Yao et~al.(2012)Yao, Fidler and Urtasun}]{urtasun}
\bibinfo{author}{Yao, J.}, \bibinfo{author}{Fidler, S.},
  \bibinfo{author}{Urtasun, R.}, \bibinfo{year}{2012}.
\newblock \bibinfo{title}{Describing the scene as a whole: Joint object
  detection, scene classification and semantic segmentation}, in:
  \bibinfo{booktitle}{Proc. IEEE Conf. Computer Vision Pattern Recog.}, pp.
  \bibinfo{pages}{702--709}.
\bibitem[{Yu and Koltun(2016)}]{crf2}
\bibinfo{author}{Yu, F.}, \bibinfo{author}{Koltun, V.}, \bibinfo{year}{2016}.
\newblock \bibinfo{title}{Multi-scale context aggregation by dilated
  convolutions}, in: \bibinfo{booktitle}{Proc. Int. Conf. Learning Represent.
  (ICLR)}.
\bibitem[{Zhao et~al.(2017)Zhao, Shi, Qi, Wang and Jia}]{psp}
\bibinfo{author}{Zhao, H.}, \bibinfo{author}{Shi, J.}, \bibinfo{author}{Qi,
  X.}, \bibinfo{author}{Wang, X.}, \bibinfo{author}{Jia, J.},
  \bibinfo{year}{2017}.
\newblock \bibinfo{title}{Pyramid scene parsing network}, in:
  \bibinfo{booktitle}{Proc. IEEE Conf. Computer Vision Pattern Recog.}, pp.
  \bibinfo{pages}{6230--6239}.
\bibitem[{Zheng et~al.(2015)Zheng, Jayasumana, Romera-Paredes, Vineet, Su, Du,
  Huang and Torr}]{crf_rcnn}
\bibinfo{author}{Zheng, S.}, \bibinfo{author}{Jayasumana, S.},
  \bibinfo{author}{Romera-Paredes, B.}, \bibinfo{author}{Vineet, V.},
  \bibinfo{author}{Su, Z.}, \bibinfo{author}{Du, D.}, \bibinfo{author}{Huang,
  C.}, \bibinfo{author}{Torr, P.H.S.}, \bibinfo{year}{2015}.
\newblock \bibinfo{title}{Conditional random fields as recurrent neural
  networks}, in: \bibinfo{booktitle}{Proc. IEEE Int. Conf. Computer Vision
  (ICCV)}, pp. \bibinfo{pages}{1529--1537}.

\end{thebibliography}

\end{document}